\documentclass[10pt,journal,compsoc]{IEEEtran}
\usepackage{amsfonts,amsmath,amssymb,amsthm,bm}
\usepackage{graphicx,psfrag,epsf,calligra}
\usepackage{enumerate}
\usepackage{adjustbox}
\usepackage{dsfont}
\usepackage{comment}
\usepackage{xcolor}
\usepackage[colorlinks=true,linkcolor=cyan,citecolor=blue]{hyperref}
\usepackage[colorinlistoftodos]{todonotes}

\newcommand{\bX}{\mathbf{X}}
\newcommand{\bD}{\mathbf{D}}

\ifCLASSOPTIONcompsoc
  \usepackage[nocompress]{cite}
\else
  \usepackage{cite}
\fi

\ifCLASSINFOpdf

\else
 
\fi

\begin{document}

\title{Data Kernel Perspective Space Performance Guarantees for Synthetic Data from Transformer Models }

\author{ Michael Browder, ~Kevin Duh, ~J. David Harris,  ~Vince Lyzinski, \\ Paul McNamee,   Youngser Park, Carey E.~Priebe, ~Peter Viechnicki
\IEEEcompsocitemizethanks{
\IEEEcompsocthanksitem Michael Browder and Vince Lyzinski are with the Department of Mathematics at the University of Maryland, College Park. E-mail: mbrowder@umd.edu, vlyzinsk@umd.edu
\IEEEcompsocthanksitem Kevin Duh, 
J.\ David Harris, 
Paul McNamee,
and Peter Viechnicki are with the Human Language Technology Center of Excellence, Johns Hopkins University. E-mail:
kevinduh@cs.jhu.edu, 
jdavidharris@gmail.com,
mcnamee@jhu.edu,
pviechnicki2@gmail.com  
\IEEEcompsocthanksitem Youngser Park
is with
the Center for Imaging Science (CIS), the Institute for Computational Medicine (ICM), and the Mathematical Institute for Data Science (MINDS), Johns Hopkins University. E-mail: parky@cis.jhu.edu 
\IEEEcompsocthanksitem Carey E.\ Priebe
is with
the Department of Applied Mathematics and Statistics (AMS), the Center for Imaging Science (CIS), and the Mathematical Institute for Data Science (MINDS), Johns Hopkins University. E-mail: cep@jhu.edu
}
}


\IEEEtitleabstractindextext{%
\begin{abstract} 
Scarcity of labeled training data remains the long pole in the tent for building performant language technology and generative AI models. Transformer models \textemdash particularly LLMs \textemdash are increasingly being used to mitigate the data scarcity problem via synthetic data generation. However, because the models are black boxes, the properties of the synthetic data are difficult to predict. In practice it is common for language technology engineers to 'fiddle' with the LLM temperature setting and hope that what comes out the other end improves the downstream model. Faced with this uncertainty, here we propose Data Kernel Perspective Space (DKPS) to provide the foundation for mathematical analysis yielding concrete statistical guarantees for the quality of the outputs of transformer models. We first show the mathematical derivation of DKPS and how it provides performance guarantees. Next we show how DKPS performance guarantees can elucidate performance of a downstream task, such as neural machine translation models or LLMs trained using Contrastive Preference Optimization (CPO). Limitations of the current work and future research are also discussed. 
\end{abstract}



}

\maketitle

\IEEEdisplaynontitleabstractindextext

%
\IEEEpeerreviewmaketitle

\IEEEraisesectionheading{\section{Introduction}\label{sec:introduction}}

\IEEEPARstart 
Neural language models have enabled unprecedented improvement in performance of language technologies for a wide range of tasks (e.g. 
automatic speech recognition (\cite{arisoy15bidir,irie19language,peng2025surveyspeechlargelanguage,asr4,asr5}), 
information retrieval ( \cite{khattab20colbert,zhu25llmforir, ir3, ir4,ir5}), particularly since the advent of transformer-based approaches \cite{vaswani2017attention,bert}. 
Transformer models have quickly achieved state-of-the-art results even on tasks of seemingly high cognitive complexity (e.g. question answering \cite{qa1,qa2,qa3}). 
In some domains such as machine translation, transformer-based large language models have become competitive with human translators in certain situations (\cite{hassan2018achievinghumanparityautomatic,mt1,mt2,mt3}).

The models achieving these impressive results however share a fundamental property with all their other neural and pre-neural forbearers - they perform worse on data that does not resemble their training data (\cite{chu-wang-2018-survey,oos1,oos2,oos3,oos4}).  
Domain sensitivity is a commonly observed phenomenon in testing and evaluation of transformer-based language technology applications, including those built from large language models (LLMs) \cite{zhang19curriculum,ds1}. Moreover, training data limitations are often easy to observe when language applications are tested against languages which are not highly resourced \cite{firat-etal-2016-zero}.

Recently a diverse set of researchers has tried to mitigate the need for expensive labeled data through the use of synthetic training data.  Synthetic data have been used to improve performance against a broad swath of tasks including speech translation \cite{bamfo-odoom-etal-2024-synthetic} and cross-language information retrieval  \cite{mayfield2023syntheticcrosslanguageinformationretrieval} to name just two. 
One could term this use of synthetic data as 'intentional synthetic data incorporation,' in contrast with  possible accidental consumption of generated data into LLMs \cite{kang2025demystifyingsyntheticdatallm}.
There seems to be a general consensus that the unintentional ingestion of data generated by an LLM into other LLMs should eventually degrade performance, \cite{collapse1,collapse2,collapse3} but estimates of how serious this problem will become and when are in their infancy.\footnote{Perhaps a leading indicator for increasingly anxiety about machine learning models consuming their own output as training data is the exponential growth in use of the term \textit{ouroboros} since 2003 \cite{googlebooks}}

One task where synthetic data have been used to particular advantage is machine translation (MT), through the mechanism of 'back translation' (BT).  In a recent extensive study of BT \cite{mcnamee-duh-2023-extensive}, incorporating synthetic data via BT is shown to provide substantial machine translation gains; specifically gains of up to thirteen BLEU points \cite{bleu} or 125\%, with the highest gains observed in low-resource languages. Beyond the broad finding that BT is most helpful for languages with the least training data, there is not yet clear consensus on where synthetic data helps MT, and where it hurts.

As more and more synthetic data is consumed by neural language models, there is therefore an urgent need to understand its quality. Most such synthetic data comes from black box models and has unrecorded provenance to boot, making it difficult to understand what effects its proliferation will have. What is needed is a framework for understanding the variability of synthetic data prior to subsequent use in transformer models. 

Here we propose 
a mathematical framework for analyzing the statistical properties of transformer model outputs. This framework permits us to quantify the distance between model inputs and outputs, even when the model weights themselves are not accessible (e.g. OpenAI's GPT family of models). We call this framework \textit{Data Kernel Perspective Space -- DKPS} \cite{acharyya2024consistent,helm2024statistical}. As we describe below, DKPS represents the mean discrepancy geometry for a collection of models from the perspective of an evaluation query set. DKPS allows us to give guarantees on the upper bound of variance in transformer-model outputs given an arbitrary input, which \textemdash we hope \textemdash will lead in turn to an understanding of the effects of synthetic data on the downstream models consuming it.

In this work, we demonstrate the application of DKPS to the (relatively simple) task of MT, which possesses certain beneficial properties for demonstrating the efficacy of our framework. We expect that DKPS will have similar efficacy for other tasks but leave demonstration  for future work.
The remainder of this article is structured as follows.  In Section \ref{sec:DKPS} we formalize DKPS and demonstrate how performance guarantees proceed from the proposed formalism. In Section \ref{sec:MT}, we show how DKPS sheds light on the properties of synthetic data in machine translation. Section \ref{sec:CPO} discusses DKPS applied in the setting of contrastive preference optimization \cite{xu2024contrastive}, and Section \ref{sec:discussions} discusses the limitations of the current work, and future research.

\section{Data Kernel Perspective Space}
\label{sec:DKPS}

Consider $n$ generative models, $f^{(1)},f^{(2)},\ldots f^{(n)}$ as random mappings $f^{(i)}:\mathcal{Q}\mapsto\mathcal{Y}$; here $\mathcal{Q}$ represents the input/query space and $\mathcal{Y}$ the output space, so that for $q\in\mathcal{Q}$ and $i\in[n]$, $f^{(i)}(q)$ is randomly distributed (according to an often unknown distribution) over $\mathcal{Y}$.
Suppose that we further have an embedding function $g:\mathcal{Y}\mapsto\mathbb{R}^p$ that provides a Euclidean representation of points in the output space.
The idea of Data Kernel Perspective Space (DKPS) is to use a collection of $m$ queries $\{q_1,q_2,\cdots,q_m\}$ to summarize and compare our models in Euclidean space as follows.
Writing the true model-query means via $\mu^{(i)}_j=\mathbb{E}(g(f^{(i)}(q_j))$, 
the $d$-dimensional DKPS representation of our models (via $g$ and $\{q_i\})$) is provided by $\Psi=\text{MDS}(\Delta)\in\mathbb{R}^{n\times d}$,
where $\Delta[i,j]=\frac{1}{m}\|\mu^{(i)}-\mu^{(j)}\|_F$, and where we use the machinery of classical Multidimensional Scaling (MDS) \cite{borg2005modern} to get the representation $\Psi$ of our collection of $n$ models (model $f^{(i)}$ represented as the $i$-th row of $\Psi$) as points in a low-dimensional Euclidean space that is amenable to subsequent inference \cite{helm2024statistical}.

To estimate the DKPS $\Psi$, we thus consider the following procedure. 
For each query $q_j$, each model produces $r$ independent random replicates $\{f^{(i)}(q_j)_k\}$; let $F_{i,j}$ denote the distribution of these replicates so that $f^{(i)}(q_j)_1,f^{(i)}(q_j)_2,\cdots,f^{(i)}(q_j)_{r}\stackrel{i.i.d.}{\sim}F_{i,j}$.
For each model $f^{(i)}$, define the summary matrix $\bX^{(i)}\in\mathbb{R}^{m\times p}$ via 
\begin{align}
    \bX^{(i)}[j,:]=\frac{1}{r}\sum_k g(f^{(i)}(q_j)_k).
\label{DKPS:ave}
\end{align}
We can then compute the estimated distances between summarized models via $\bD\in\mathbb{R}^{n\times n}$, where $\bD[i,j]=\frac{1}{m}\|\bX^{(i)}-\bX^{(j)}\|_F$.
As $\bD$ is a Euclidean distance matrix (up to scaling), we can compute $\widehat{\Psi}=\text{MDS}(\bD)\in\mathbb{R}^{n\times d}$ so that for each $i\in[n]$, 
$\widehat{\Psi}[i,:]$ represents the estimated MDS summarization (provided via $g$ and $\{q_i\}$) of $f^{(i)}$ in $\widehat{\Psi}$.
As $r\rightarrow\infty$, $\bD$ converges $\Delta$, and in \cite{acharyya2024consistent} it is shown that $\widehat{\Psi}$ provides a consistent estimate of $\Psi$ under mild conditions.

\section{DKPS applied to Machine Translation}
\label{sec:MT}

To illustrate the utility of DKPS, we consider the following illustrative example.
We consider a data set consisting of $n_I=1000$ English sentences, each paired with a human-translated Zulu counterpart.
Zulu is a relatively low-resource language \cite{zulu}, and thus provides a good setting to explore the utility of synthetic data augmentation, here provided by a bespoke English-Zulu MT model built within the Sockeye Neural Machine Translation platform \cite{hieber2017sockeye}.
Notably, these 1000 sentence pairs are from the training set used for our en-zu MT model; we will further consider another collection of 1000 sentences of English---with human Zulu translations---from the en-zu model validation set to explore in- versus out-of-sample (OOS) distribution differences below; we note here that the MT model did not translate one of the OOS sentences (no output is produced), so the OOS data set below will be $n_O=999$ sentence pairs.
Understanding the potentially different statistical properties of the human versus synthetic datasets both in- and -out-of-sample is of paramount import, and DKPS provides a natural pathway for such exploration.

\begin{figure}[t!]
  \centering
\includegraphics[width=0.45\textwidth]{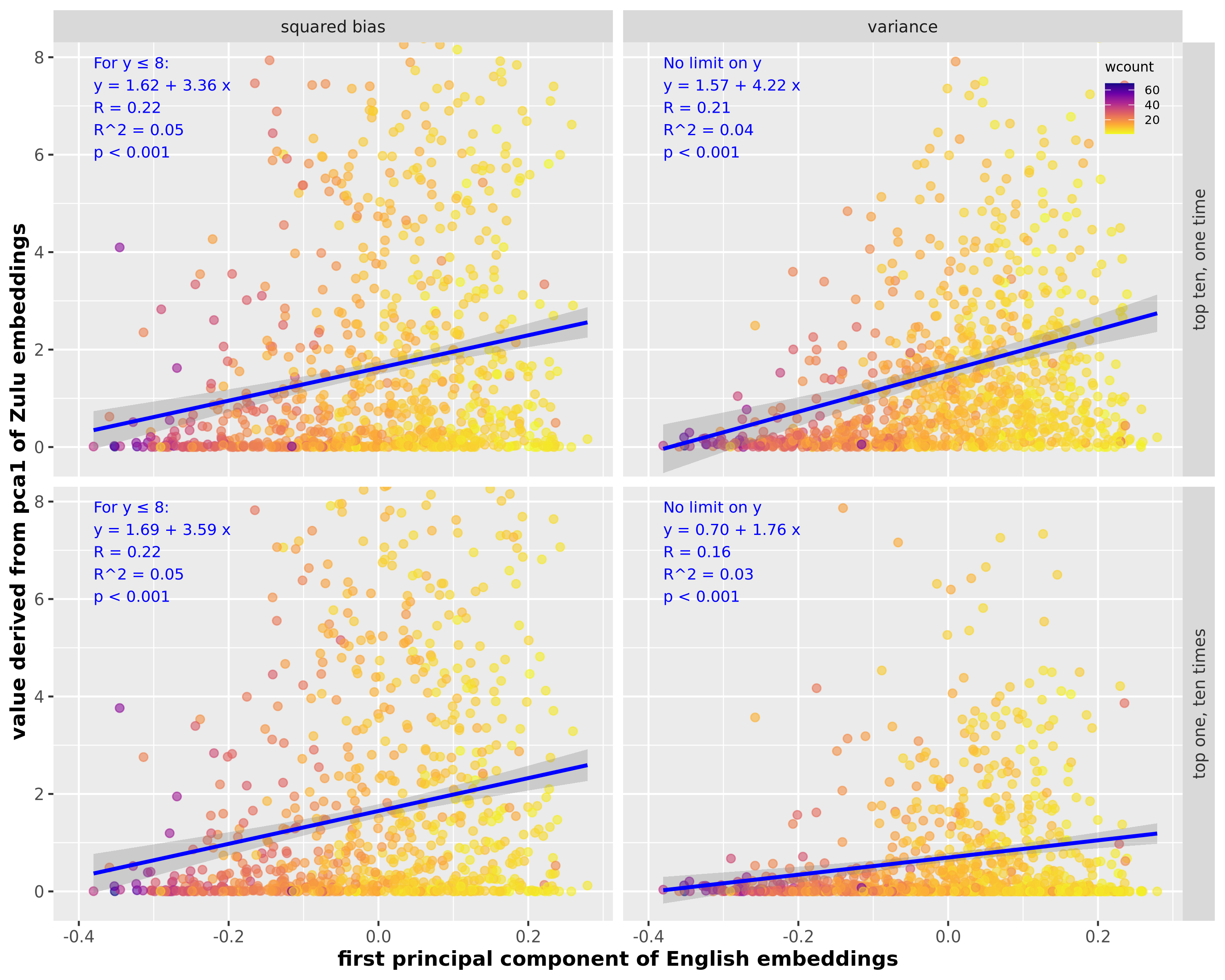}
  \caption{
For temperature $T=0.6$, we plot: (Left) Squared bias for the 10 batched synthetic translations with respect to the human translation; (Right) Estimated variance for the 10 synthetic translations. 
The upper row is the batched in-sample translations and the bottom the sequential in-sample translations.
Points are colored by the number of words of English sentences.
}
\label{fig:bias_variance_in}
\end{figure}

\subsection{Estimation fidelity}
\label{sec:est}

For each of the in-sample and out-of-sample (OOS) English-Zulu human translation pairs, we first use standard dimension reduction methods to explore the statistical properties (bias and variance) of the en-zu machine translations as estimators of the human translation.
For each of the English sentences, we consider two scenarios:
\begin{itemize}
    \item[i.] We query the MT model $t=10$ times (sequentially), and choose the top provided translation for each of the 10 queries.  We can view these queries as 10 i.i.d. translations from the model.  This is done both for the in-sample and OOS data.
    From this we can explore in-sample versus out-of-sample properties.
    \item[ii.] We use the top $t=10$ ranked translations provided by a single MT query of the in-sample data.  From this, we can explore the distributional difference between the top-produced machine translation and the lower-ranked translations. 
\end{itemize}
We then use LASER3 \cite{LASER,LASER3} representations (at temperature $T=0.2, 0.6, 1$; results are shown for $T=0.6$ ; for other temperatures see Figure \ref{fig:bias_variance} in Appendix \ref{sec:figs}) to embed the English sentence collection, the human translations, and the synthetic translation sets into $\mathbb{R}^{1024}$.
Given the $t=10$ sample size in each case, this embedding is prohibitively high dimensional for subsequent inference---here mean and variance estimation, with more complex inference tasks  most likely requiring higher sample size and dimension---and we consider using principal component analysis (PCA) \cite{abdi2010principal} to further reduce the dimensionality of the sentence embeddings.
Based on the PCA SCREE plots (see Figure \ref{fig:scree} in Appendix \ref{sec:figs}) of the in-sample and OOS English embeddings and human-translated Zulu embeddings, we infer that $p$ here equal to 1,2, 3 or 4 is a suitable dimension for exploring the statistical properties of the synthetic translations.
We show below results for $p=1$, but results are similar for $p=2,3,4$.

\begin{figure}[t!]
  \centering
\includegraphics[width=0.45\textwidth]{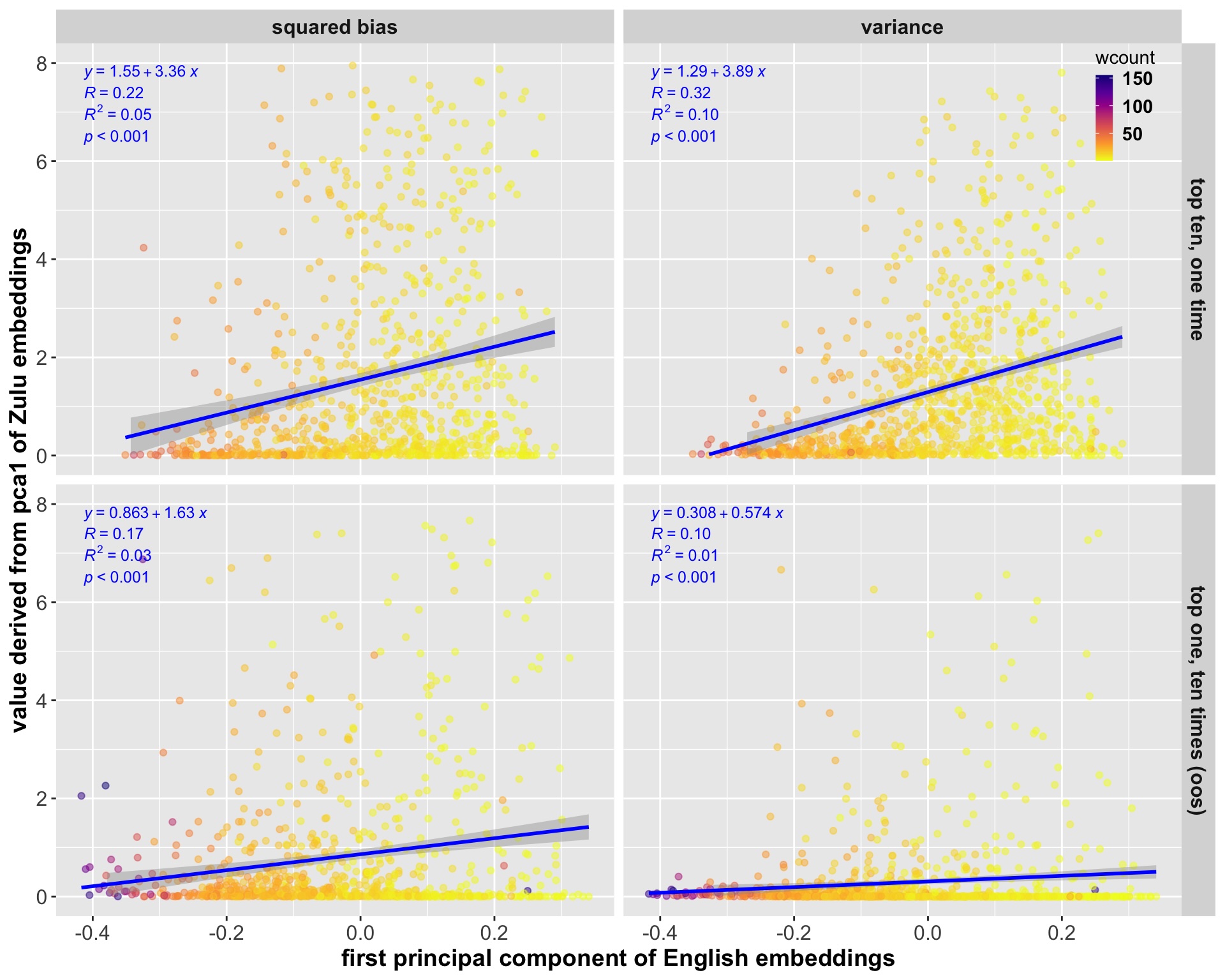}
  \caption{
For temperature $T=0.6$, we plot: (Left) Squared bias for the 10 batched synthetic translations with respect to the human translation; (Right) Estimated variance for the 10 synthetic translations. 
The upper row is the batched in-sample translations and the bottom the sequential OOS translations.
Points are colored by the number of words of English sentences.
}
\label{fig:bias_variance_out}
\end{figure}

Viewing the human translation embedding as the ideal, we can estimate the bias and variance of the synthetic Zulu translations as follows:
First, let $z_{ik}$ be the (1-d) PCA embedding of the $k$-th synthetic translation of the $i$-th English sentence, and let $z_i$ be the (1-d) PCA embedding of the human translation of the $i$-th English sentence.
We can then estimate the squared bias and variance via (where $\bar{z}_{i\bullet}=\sum_k z_{ik}/t$)
$
\text{bias}_i=\left(\bar{z}_{i\bullet} -z_i\right)^2,$ and $\text{Var}_i=\sum_k\left(z_{ik}-\bar{z}_{i\bullet}\right)^2/(t-1).$ 
Results are plotted in Figures \ref{fig:bias_variance_in} and  \ref{fig:bias_variance_out}.
From the figures, we see that even in this 1-dimensional setting both squared bias and variance vary significantly and predictably according to the word count of the sentences; in all settings, shorter sentences produce more bias and more variance in the embedded space (variance also increases predictably with temperature; see Figure \ref{fig:bias_variance}).
In the in-sample setting plotted in Figure \ref{fig:bias_variance_in} (where the $n_I=1000$ English sentences are embedded without the OOS data), the bias is relatively constant between the sequential and batched translations, though the variance of the batched translations is higher as expected.
This is of note, as the batched translations are returning sub-optimal (i.e., ranked lower than 1) translations which though more variable, do not add additional bias into the system; we explore this further via the DKPS machinery in the following section.

Comparing the in- and OOS settings in Figure \ref{fig:bias_variance_out} (here the $n_I+n_O=1999$ English sentences are jointly used to compute the PCA; the in-sample and out-of-sample translations are jointly used for the Zulu PCA as well), we see that both bias and variance for the OOS translations is lower than the in-sample sequential and batched translation counterparts (notably, the variance is markedly lower).
Note that these are difference datasets being compared; the OOS data contains longer sentences whose translations are less variable in general. Moreover, heuristics are applied during MT model training to prevent overfitting to the training data.
In all cases, debiasing the in-sample and OOS synthetic data can be inferentially beneficial \cite{decruyenaere2024debiasing}, and the predictable nature of the bias/variance (as a function of the first PCA component and the sentence length) lends itself to a simple debiasing procedure.

\begin{figure*}[t!]
\centering
\includegraphics[width=0.8\textwidth]{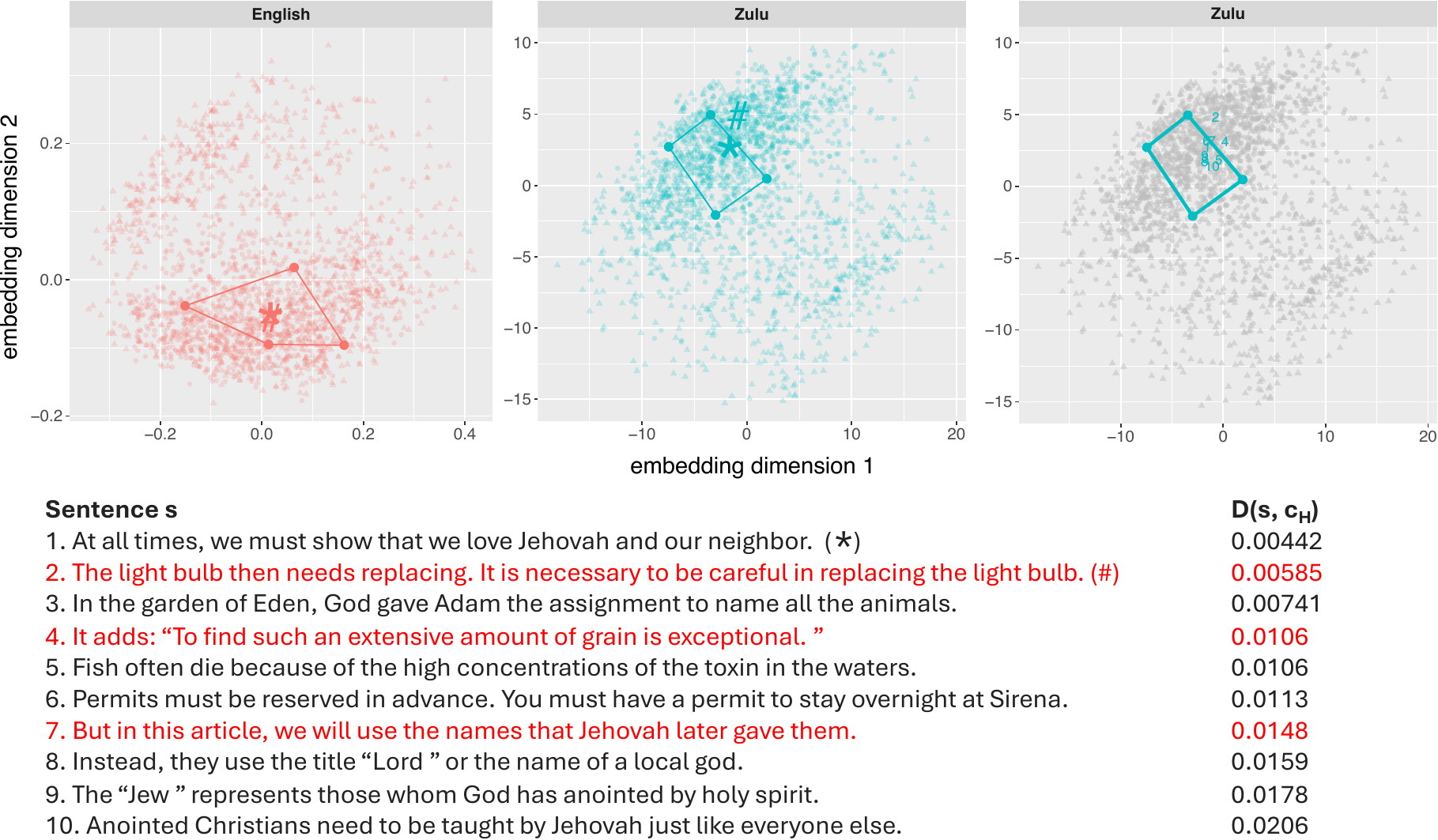}
\caption{Convex hull of the points $\mathcal{C}_E$ (L) and $\mathcal{C}_Z$ (M and R).  Below we show the ten closest OOS sentences to the mean of the English convex hull (the closest is denoted via an $*$, the second via a $\#$).  Sentences in red (resp., black) font have their translations mapped outside (resp., inside) the convex hull of $\mathcal{C}_Z$.
In the right panel, the translations of the ten sentences are plotted and numbered 1--10.
}
\label{fig:zendian}
\end{figure*}

\begin{figure}[t!]
\centering
\includegraphics[width=0.45\textwidth]{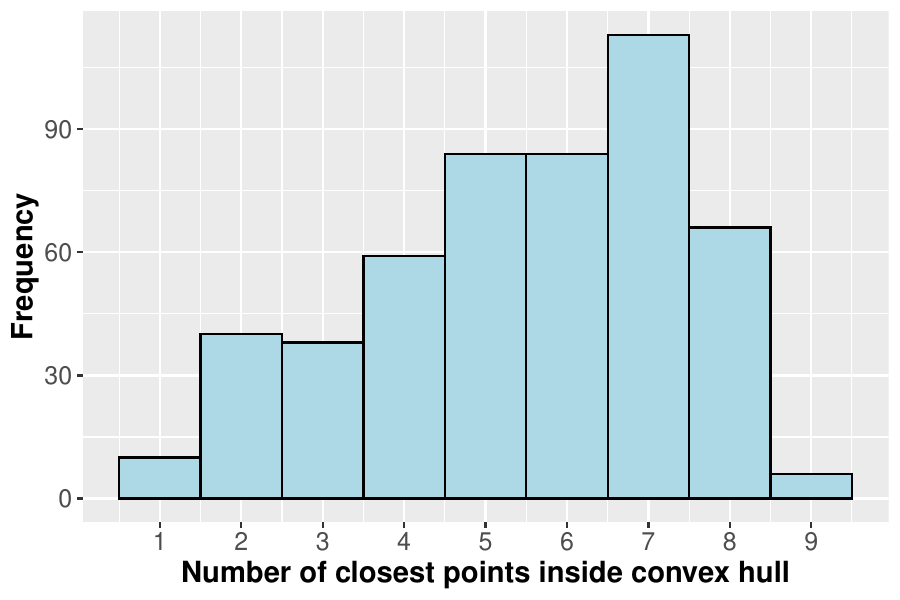}
\caption{Choosing 500 random sets $\mathcal{C}_E$, we plot the histogram for the number of the ten closest sentences to $\mathcal{HC}_E$ whose translations fall into $\mathcal{HC}_Z$.
The mean is $5.47\pm 2$.
}
\label{fig:zendian2}
\end{figure}

\subsection{Moving beyond estimation}
\label{sec:beyond}

Understanding the properties of synthetic data beyond estimation necessitates a deeper understanding of the geometry and generalizability of the ``translation$\circ$embedding" pipeline.
To illustrate the complexities here, we consider the following illustrative experiment.
We consider the 2-d PCA embeddings of the 1024-dimensional English sentence embeddings (via LASER3); here we consider all 1999 sentences (1000 in-sample and 999 OOS).
For each in-sample sentence, we consider the 2-d PCA of the LASER3 embedding of the human (i.e., ``ground truth'') Zulu translation; for each OOS sentence, we consider the 2-d PCA of the embedding derived via LASER3$\circ$MT
using the top ranked machine translation.

If the geometry of the sentences in the embedding space is preserved by the translation platform (i.e., preserved in the synthetically generated data), 
then we would expect that if an OOS English sentence is close to an in-sample English sentence, then the machine translation of the OOS should be close to the human translation of the in-sample sentence.
To understand if this is the case, we consider 4 randomly chosen in-sample English sentences (call this set $\mathcal{C}_E$), and construct their convex hull in PCA space (in red in Figure \ref{fig:zendian}) and the analogous convex hull $\mathcal{HC}_E$ of the human translations (in blue in Figure \ref{fig:zendian}; call these sentences $\mathcal{C}_Z$ and their hull $\mathcal{HC}_Z$).
OOS points in the convex hull are, in a sense, ``explained'' by the sentences in $\mathcal{C}_E$, and it is reasonable to expect their synthetic translation is in $\mathcal{HC}_Z$.
For synthetic data augmentation, we may wish to use only candidate English sentences $e'$ in the convex hull of $\mathcal{C}_E$ such that the Zulu translation is in the convex hull  $\mathcal{C}_Z$ with high probability.

In Figure \ref{fig:zendian}, we show the ten closest OOS sentences to the mean of $\mathcal{HC}_E$ (the closest is denoted via an $*$, the second via a $\#$).  Sentences in red (resp., black) font have their translations mapped outside (resp., inside) of $\mathcal{HC}_Z$.
We note here that most of these ten sentences are religious in nature.  Notably, sentences $\{2,4,5,6\}$ are not, and 
$\{2,4\}$ have their translations mapped outside of $\mathcal{HC}_Z$, and the translations of $\{5,6\}$ are near the boundary of $\mathcal{HC}_Z$.
All of the religious themed sentences have translations mapped inside of $\mathcal{HC}_Z$ except for sentence 7 (though it is mapped near the border of $\mathcal{HC}_Z$).
While this experiment is illustrative, to understand the broader phenomena we repeat the above with 500 randomly chosen sets $\mathcal{C}_E$, and in Figure \ref{fig:zendian2}, we plot the histogram for the number of the ten closest sentences to the center of $\mathcal{HC}_E$ whose translations fall into $\mathcal{HC}_Z$.
From the figure, we see that our illustrative example is emblematic of a broader phenomena.
Indeed, the process of embedding and translation, though difficult to predict in general, nonetheless at least partially preserves the geometry and syntax of the OOS synthetic data.

\subsection{DKPS for comparing Batch versus sequential (in-sample)}
\label{sec:bVs}

\begin{figure}[t!]
  \centering
\includegraphics[width=0.47\textwidth]{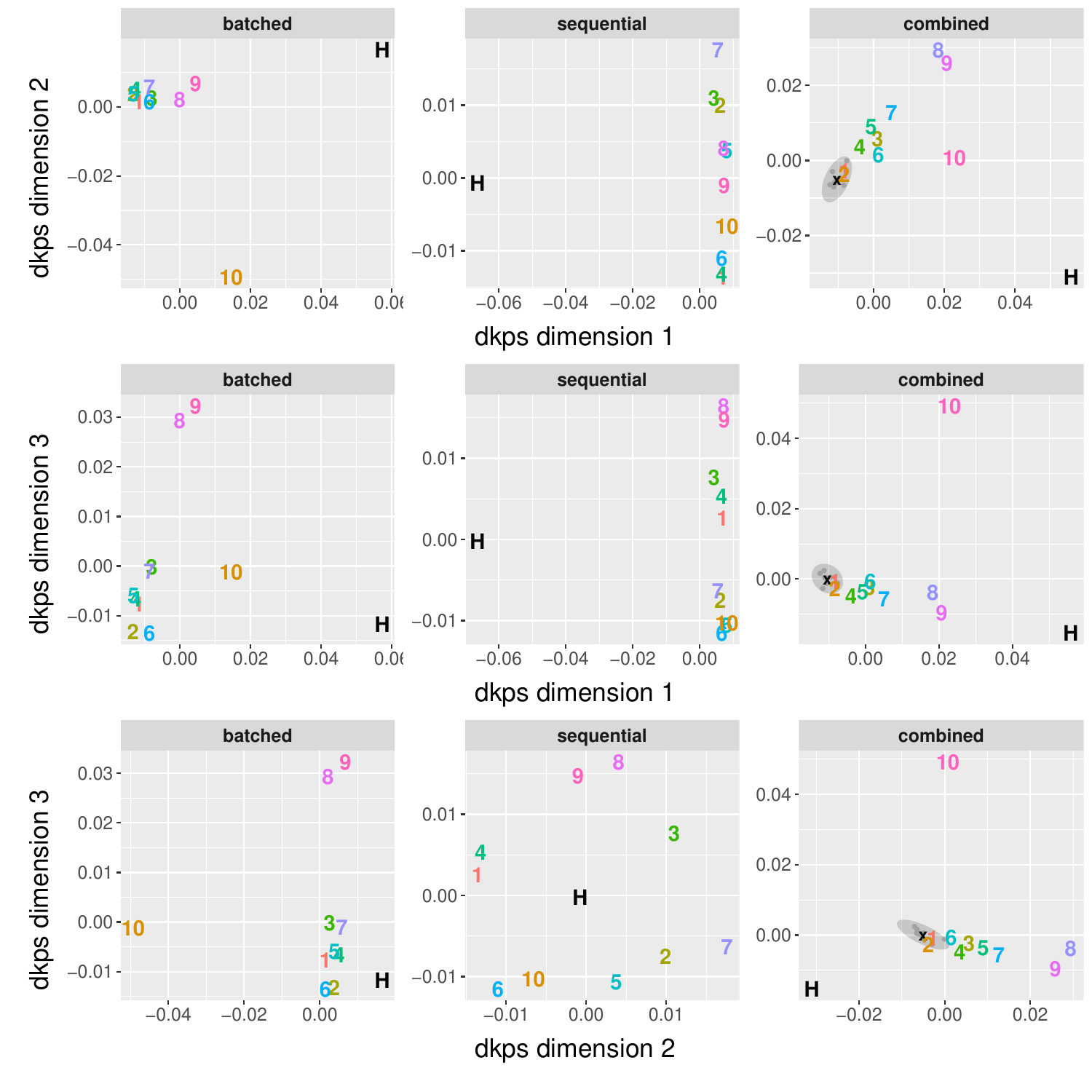}
  \caption{
Using MDS to embed the distance matrices in Figure \ref{fig:Din} into an appropriate $\mathbb{R}^d$, we show the pairs plot of the first three DKPS dimensions in each case.
In the first two columns, the plot the batched and sequential models and the 10 translations are represented via numerals 1-10 (in the batched setting numbered according to their Sockeye rank) with the human translation model provided by the point labeled $H$.
In the third column, the sequential translation models are provided by the black dots with a fitted Gaussian distribution in gray and mean show via the $x$.
The batched are again numbered 1-10 (according to Sockeye rank).}
\label{fig:dkps}
\end{figure}

While the in-sample sequential translations can be viewed as i.i.d., the batch translations (one query returning the top $t=10$ system outputs) are not.
The batch translations return lower ranked (by the MT system) translations which may allow for better exploration of the translation space, at the expense of injecting additional variance (and not bias) into the system.
DKPS provides a tool to interrogate this further, giving us means to better understand the form and structure of the model parameters elucidated above.  
To this end, we first treat the 1000 in-sample English sentences as the DKPS queries $\{q_i\}$.
We consider using DKPS to summarize the following models
$$
\underbrace{f^{(1)}}_{\substack{\text{human}\\\text{ translation}}},
\underbrace{f^{(2)},\cdots,f^{(11)}}_{\substack{\text{10 i.i.d. Sockeye best}\\\text{translations (sequential)}}}
,\underbrace{f^{(12)},\cdot,f^{(21)}}_{\substack{\text{the 10 ranked Sockeye}\\\text{translations (batch)}}}.
$$
As each query in the batch setting produces one 1 independent replicate (i.e., $r=1$ in the DKPS notation) per query per model ($f^{(11+i)}$ modeling the $i$th ranked translation), we choose not to average over the $t=10$ sequential (i.i.d.) replicates as in Eq. \ref{DKPS:ave}---this is to produce and compare models with commensurate training set sizes.
Rather, we consider each sequential query as producing one independent replicate for each of $t=10$ identically distributed models $f^{(i)}$ for $i=2,3,\ldots,11$; in the notation of DKPS we model this as querying each of the 21 models $r=1$ time.
Working in 1-dimensional PCA space, we then have that each $\mathbf{X}^{(i)}$ is in $\mathbb{R}^{1000\times 1}$.
We then compute (plotted in Figure \ref{fig:Din} in Appendix \ref{sec:figs}) the distance matrices in 1-dimensional PCA space and the MDS SCREE plots between: 
\begin{itemize}
    \item[i.] the human translation and the batched translations ($\mathbf{X}^{(1)},\mathbf{X}^{(12)},\cdots\mathbf{X}^{(21)}$); here only the human and batched translations are used to compute the PCA embedding
    \item[ii.] the human translation and  sequential translations ($\mathbf{X}^{(1)},\mathbf{X}^{(2)},\cdots\mathbf{X}^{(11)}$); here only the human and sequential translations are used to compute the PCA embedding
    \item[iii.] the human translation and the combined batched and sequential translations
(all $\mathbf{X}^{(i)}$); here all translations are used to compute the PCA embedding.
\end{itemize}
\noindent The MDS geometry dimension varies significantly across the cases:
The batched translations are noisier, have a higher DKPS dimension, and vary more within sample and when compared to the human translation.
Using MDS to embed these collections into an appropriate $\mathbb{R}^d$, we plot in Figure \ref{fig:dkps} the pairs-plot of the first three DKPS dimensions in each setting.
In the first two columns, we plot the batched and sequential models and the 10 models are represented via numerals 1-10 (in the batched setting numbered according to Sockeye rank) with the human translation model provided by the point labeled $H$.
In the third column, the sequential models are provided by the black dots with a fitted Gaussian distribution in gray and mean shown via the $x$.
The batched are again numbered 1-10 (according to their MT system rank).

From Figure \ref{fig:dkps}, we see that in the sequential model setting, the bias is captured in the first principle component, with the second and third components capturing the variability in the sample.
In the batched models, bias and variance are present in all three top DKPS dimensions.  
If the first dimension here is capturing bias and the second and third variability, then there is a small difference in the bias among the 10 batch models, with the lower ranked models being more variable presenting itself in higher dimensions in DKPS space.
When combined, we see a blending of the two geometries.
Importantly, if we denote the true distribution of the sequential models via $F$, there is a meaningful difference between the batched model distributions and $F$ here, with distance from $F$ roughly correlating to the translation rank of the batched model.

\begin{figure}[t!]
  \centering
\includegraphics[width=0.47\textwidth]{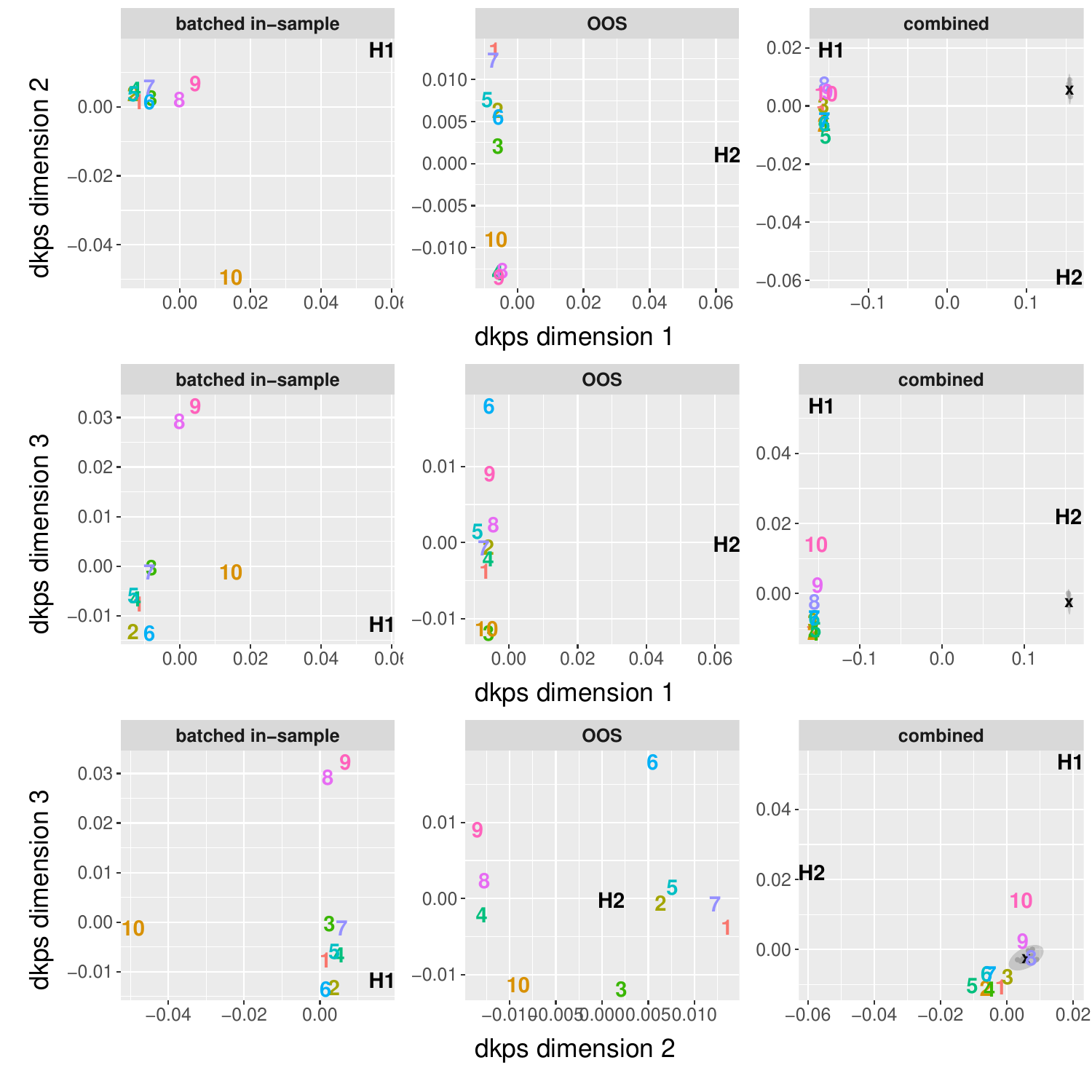}
  \caption{
Using MDS to embed the distance matrices in Figure \ref{fig:Dout} into an appropriate $\mathbb{R}^d$, we show the pairs plot of the first three DKPS dimensions in each case.
In the first two columns, the plot the batched (in-sample)  and sequential OOS models and the 10 models are represented via numerals 1-10 (in the batched setting numbered according to their Sockeye rank) with the human translation models provided by the point labeled $H1$ and $H2$.
In the third column, the sequential OOS translations are provided by the black dots with a fitted Gaussian distribution in gray and mean show via the $x$.
The batched are again numbered 1-10 (according to Sockeye rank).}
\label{fig:dkps2}
\end{figure}

\subsection{DKPS for comparing batch versus sequential (in-sample versus OOS)}
\label{sec:bVsO}

We repeat the illustrative analysis of Section \ref{sec:bVs} with the in-sample batch models and the out-of-sample sequential models: 
here DKPS summarizes the models
$$
\underbrace{f^{(1)}}_{\substack{\text{in-sample human}\\\text{ translation (H1)}}},
\underbrace{f^{(2)}}_{\substack{\text{OOS human}\\\text{ translation (H2)}}}
\underbrace{f^{(3)},\cdots,f^{(12)}}_{\substack{\text{10 i.i.d. OOS Sockeye}\\\text{ best translations}\\\text{(sequential)}}}
,\underbrace{f^{(13)},\cdot,f^{(22)}}_{\substack{\text{10 ranked Sockeye}\\\text{in-sample translations}\\\text{(batch)}}}
$$
As above, each $\mathbf{X}^{(i)}$ is in $\mathbb{R}^{1000\times 1}$, and we compute the relevant distance matrices, and again we see that the MDS geometry dimension varies significantly across the cases (see Figure \ref{fig:Dout} in Appendix \ref{sec:figs}).
The batched models are noisier, have a higher DKPS dimension, and vary more within sample and when compared to the human translation model.
In Figure \ref{fig:dkps2}, we plot the first three DKPS dimensions in each of the three settings.
As before, we see that in the sequential OOS translation model setting, the bias is captured entirely in the first principle component, with the second and third components capturing the variability in the sample.
In the batched translation models, bias and variance are present in all three top DKPS dimensions as before.  
When combined, we see a fascinating blending of the two geometries.
The two human translations are markedly different, and this causes a de-biasing in the batched and OOS translations in the first DKPS dimension, as well as reduced variability of the translations across all three dimensions.
When compared to the in-sample setting where the data were more commensurate, DKPS here elucidates the capacity of the translation$\circ$embedding pipeline combining different synthetic data sets (with markedly different properties) in a way that jointly denoises the disparate geometries.
\section{DKPS for CPO}
\label{sec:CPO}
In the above analysis, the estimation of bias and variance and the model comparison are (at least implicitly) based on a maximum likelihood framework for model estimation.
To show the flexibility of the DKPS framework, we also consider applying the DKPS machinery in the context of Contrastive Preference Optimization (CPO) \cite{xu2024contrastive}, a leading approach for the post-training of a model.
CPO is an approach that builds on its predecessor, Direct Preference Optimization (DPO) \cite{dpo}, which has shown to be effective and widely used in a variety of experiments involving large language models such as text summarization, dialogue, and sentiment analysis.  
Moreover, CPO has demonstrated state of the art performance in a number of machine translation tasks.
The effectiveness of CPO flows from the design of its loss function which builds upon the DPO objective.
Like DPO, CPO is a reinforcement learning framework:
triplets of data $(x,y_w,y_l)$---here $x$ is the data to be translated, $y_w$ a preferred translation and $y_l$ a dispreferred alternative---are used to optimize the following objective function over models $\pi_\theta$:
\[
\mathcal{L}_{\text{CPO}}(\theta)
  = \mathcal{L}_{\text{prefer}}(\theta)
  + \mathcal{L}_{\text{NLL}}(\theta)
\]
where
\begingroup
\[
\begin{aligned}
\mathcal{L}_{\text{prefer}}(\theta)
&= - \mathbb{E}_{(x, y_{w}, y_{l})\sim P_{D}}
   \log \alpha \Bigl[\beta\bigl(\log \pi_{\theta}(y_{w}\mid x) \\
&\quad\quad
   - \log \pi_{\theta}(y_{l}\mid x)\bigr)\Bigr], \\[2pt]
\mathcal{L}_{\text{NLL}}(\theta)
&= -\mathbb{E}_{(x, y_{w})\sim P_{D}}\Bigl[ \log \pi_{\theta}(y_{w}\mid x)\Bigr],\\[2pt]
\alpha
&=\frac{1}{1+e^{- \beta \Delta}};\quad
\Delta=\log \pi_{\theta}(y_{w}\mid x)\! -\! \log \pi_{\theta} (y_{l}\mid x)
\end{aligned}
\]
\endgroup
Here we model the data (the sentences in 1-d and 3-d PCA space) using a Gaussian mixture model framework with equal weights via 
$\pi_\theta(y|x)\propto\phi_w(y|x)^z\phi_l(y|x)^{(1-z)}$---here $\phi_w$, $\phi_l$ are the associated Gaussian densities, and $z=1$ if $y$ is from the preferred model $\phi_w$; moreover, we will assume the $z$'s are known in the training framework.
In this illustrative example, we will consider the sequential data to be the preferred model (i.e., drawn from $\phi_w$), and the batch the dispreferred (i.e., drawn from $\phi_l$). 
Note that the sequential data can be viewed as i.i.d. drawn from $\phi_w$ while the batch data is not i.i.d.; rather, here we posit the batch data as identically distributed from $\phi_l$ with \emph{dependence}.
When optimizing the above function, we fit a model corresponding to each of the 1000 in-sample sentences using the $20$ sequential and batched translations, and we optimize $\mathcal{L}_{\text{CPO}}(\theta|z)$ so that we assume the latent class membership variables are known a priori. 

\begin{figure}[t!]
  \centering
\includegraphics[width=0.45\textwidth]{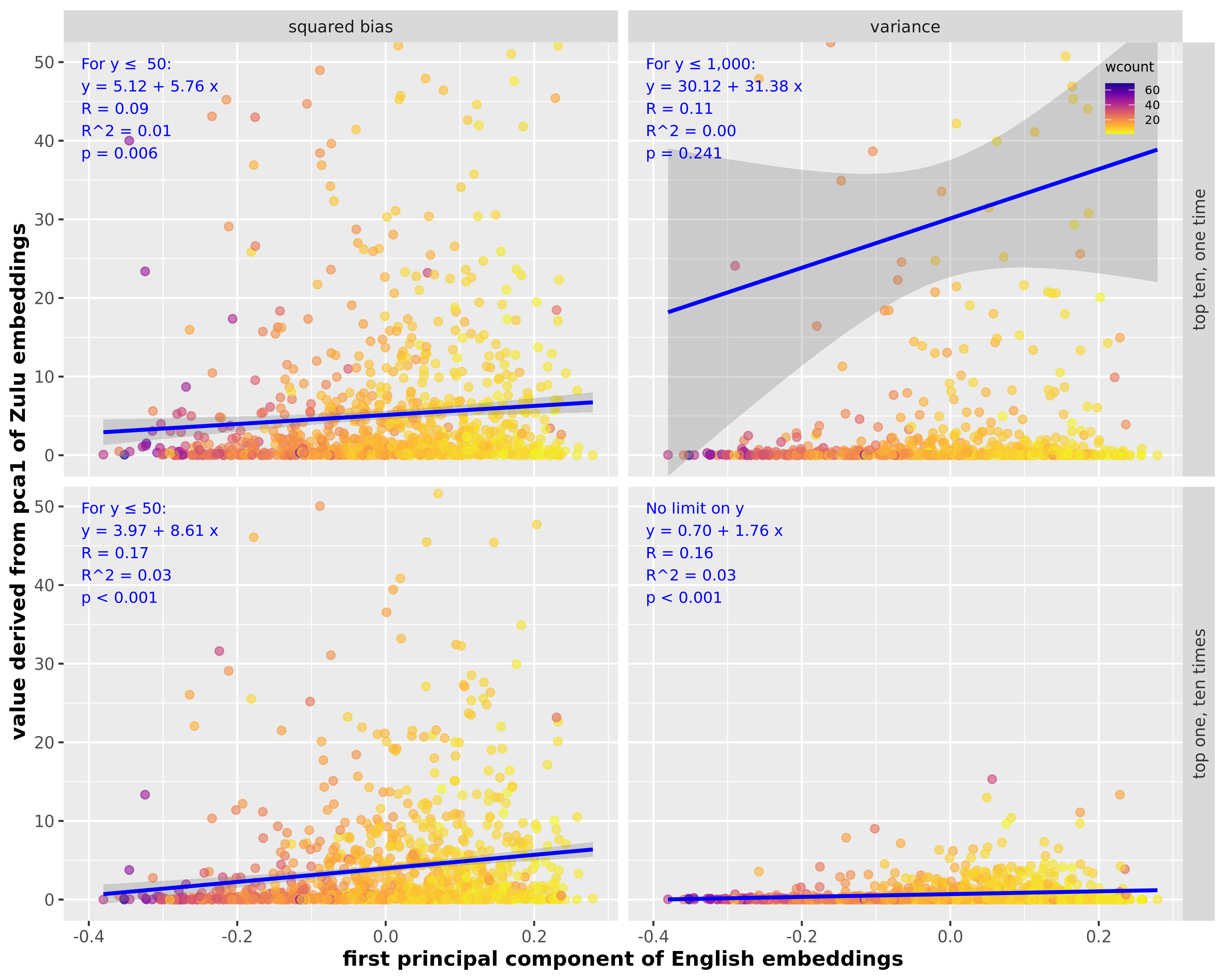}
  \caption{
  For temperature $T=0.6$, we use CPO to estimate squared bias with respect to the human translation (L) and variance (R) of the preferred (here sequential) and dispreferred (here batch) synthetic data models.
The upper row is the batched in-sample translations and the bottom the sequential in-sample translations.
Points are colored by the number of words of English sentences.}
\label{fig:CPO_bias_variance}
\end{figure}

As above, we can estimate the statistical properties of bias and variance in this CPO framework.  In Figure \ref{fig:CPO_bias_variance}, we plot the bias and variance of the CPO estimator in 1-d PCA space (to match Figure \ref{fig:bias_variance_in}).
We see that the effect of the reinforcement learning framework is to dampen the variance of the preferred sequential translations at the expense of significantly higher overall variance for the batch translations.  Interestingly, bias seems to be relatively unaffected, and, as before, both bias and variance vary predictably with sentence length.
We observe that compared to the MLE case, CPO inflates $\Sigma_{l}$, which is expected as  
this makes $\log \pi_{\theta}(y_{l}\mid x)$ small and the magnitude of $\mathcal{L}_{\text{prefer}}$ larger as desired.  
Interestingly, we found that our parameter estimates for $\phi_w$ were not identical to but closely matched those from the MLE optimization.  
This is not entirely surprising, as for $\beta$ =  0, $\theta_{w_{\text{CPO}}}$ is identical to $\theta_{w_{\text{MLE}}}$. 

\begin{figure}[t!]
    \centering
    \includegraphics[width=1.0\linewidth]{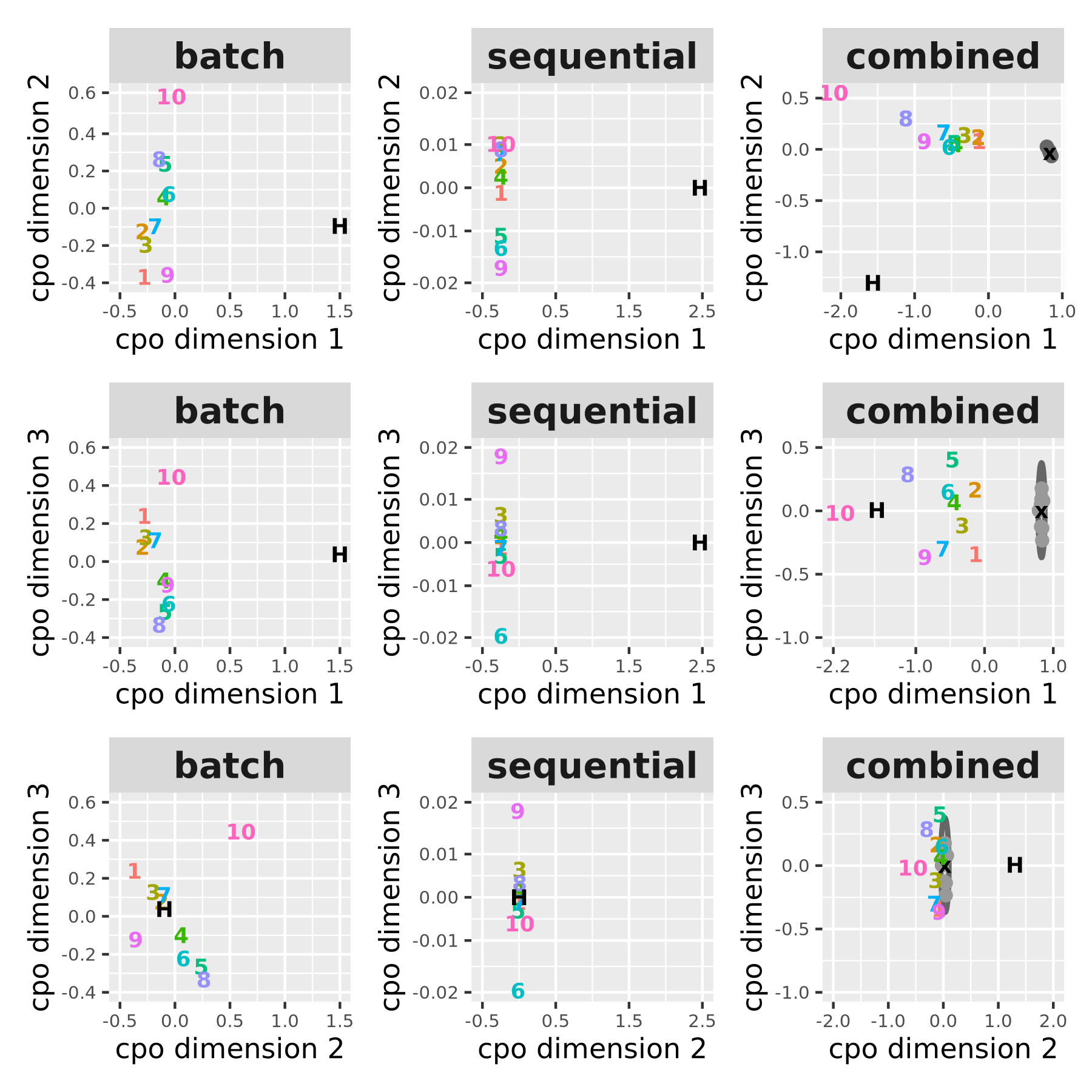}
    \caption{
Using MDS to embed the Mahalanobis distance matrices in the CPO setting of the 1-d PCA embeddings:
In the first two columns, the plot the batched and sequential translations and the 10 translations are represented via numerals 1-10 (in the batched setting numbered according to their Sockeye rank) with the human translation provided by the point labeled $H$.
In the third column, the sequential i.i.d. translations are provided by the black dots with a fitted Gaussian distribution in gray and mean show via the $x$.
The batched are again numbered 1-10 (according to Sockeye rank).  
Note the differing scales on the x- and y-axes.
    }
    \label{fig:dkpscpo1}
\end{figure}

\begin{figure}[t!]
    \centering
    \includegraphics[width=1.0\linewidth]{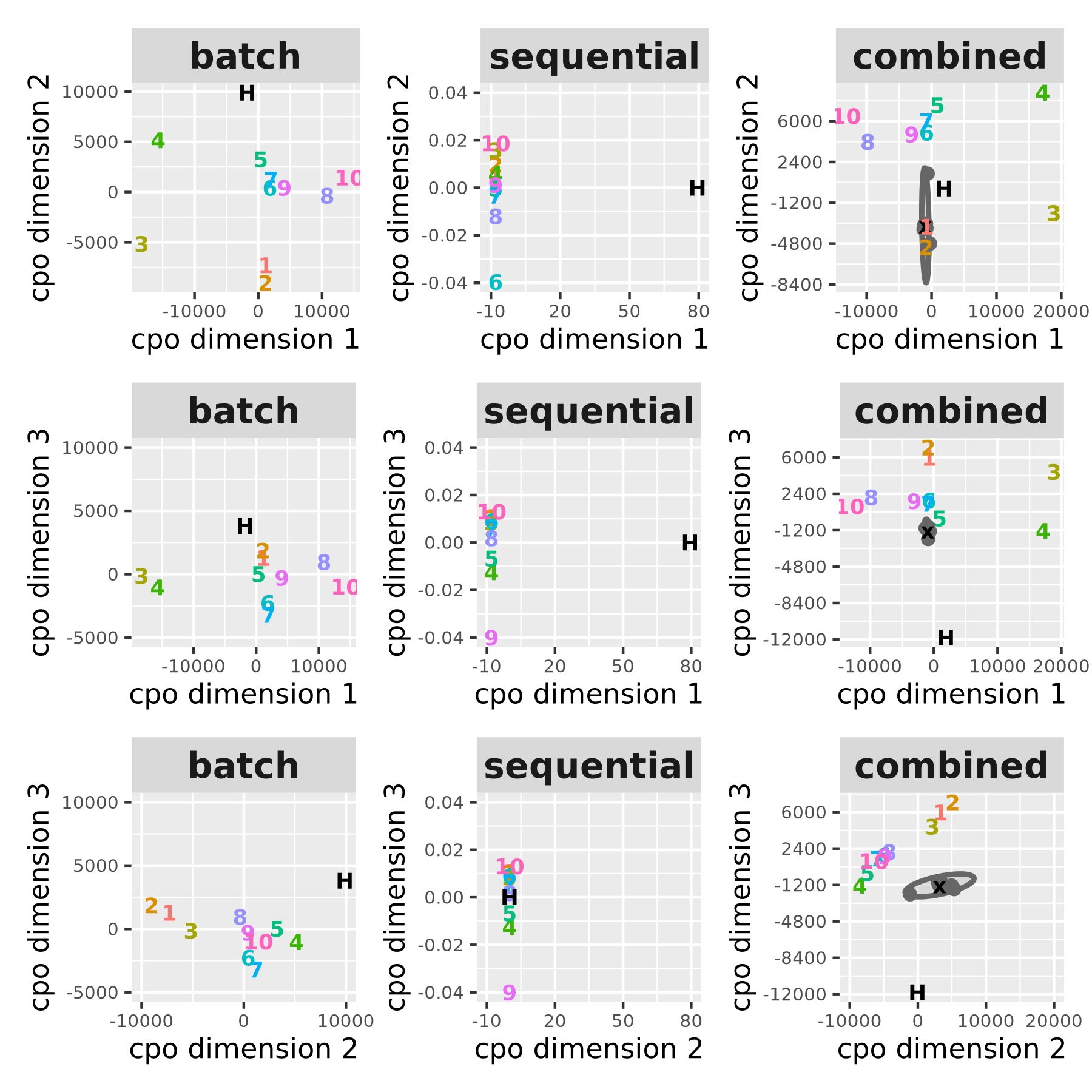}
    \caption{Using MDS to embed the Mahalanobis distance matrices in the CPO setting of the 3-d PCA embeddings:
In the first two columns, we plot the batched and sequential translations and the 10 translations are represented via numerals 1-10 (in the batched setting numbered according to their Sockeye rank) with the human translation provided by the point labeled $H$.
In the third column, the sequential i.i.d. translations are provided by the black dots with a fitted Gaussian distribution in gray and mean show via the $x$.
The batched are again numbered 1-10 (according to Sockeye rank).
Note the differing scales on the x- and y-axes.}
    \label{fig:dkpscpo3}
\end{figure}

As in the maximum likelihood setting of Section \ref{sec:bVs}, we applied DKPS to further elucidate our results from CPO model fitting.  
Due to the very different covariance structures between the batch and sequential models here, we use Mahalanobis distance as the distance metric for DKPS (see Appendix \ref{sec:dkps} for detail on the computation of the distance matrices).  
As in Figure \ref{fig:dkps}, we consider the 3-d DKPS of the 1-d PCA  embedding in Figure \ref{fig:dkpscpo1} (denote this setting DKPS$_{1\rightarrow 3}$); we also consider the analogous DKPS of the 3-d PCA embedding in Figure \ref{fig:dkpscpo3} (denote this setting DKPS$_{3\rightarrow 3}$). 
In both cases, DKPS gives insight into the \emph{structure} of the bias and variance of the models, and this structure varies meaningfully across dimensions; for example, for sequential DKPS$_{1\rightarrow 3}$ and DKPS$_{3\rightarrow 3}$, the structured bias appears in the first dimension and not as much in the other dimensions and the variance appears in components 2 and 3 and less in component 1.  This is also what we observed in Figure \ref{fig:dkps} of the MLE case.  This is not a total surprise, as we have observed above the similarity in $\hat{\theta}_{w_{\text{MLE}}}$ and $\hat{\theta}_{w_{\text{CPO}}}$.  

As we noted above, CPO tends to inflate $\Sigma_{l}$ and even in the context of Mahalanobis distance, larger variance generally appears in batch compared to sequential (note the differing scales on the x-and y-axes).  
Indeed, in the combined setting, we also see that the preferred sequential translations have less 
variability than the dispreferred batch translations. 
In DKPS$_{1\rightarrow 3}$, there appears more bias in sequential compared to batch for component 1.  In this sense, the results exemplify the classic  inverse relationship between bias and variance.  
This trend does not hold in DKPS$_{3\rightarrow 3}$, as the full 3-d batch models are more biased/variable than the full 3-d sequential models. 
An interesting phenomenon is observed in the combined setting in DKPS$_{3\rightarrow 3}$.  We see that the sequential models, though still less variable and biased than the batched counterparts, have dramatically increased bias/variance as compared to the sequential models considered alone.
This demonstrates a potential pitfall when incorporating synthetic data into translation models, as the higher bias/variance of the dispreferred models bleeds into the preferred models.
The preferred data becomes more contaminated.
This trend is also observed, though less pronouncedly, in DKPS$_{1\rightarrow 3}$.
While we could see some of these insights concerning bias and variance in the figures above, DKPS gives the additional insight into the structure of the bias and variance and how these key properties vary across MDS dimensions.

\section{Discussion and Limitations}
\label{sec:discussions}

Many learning models are black boxes, and both the properties of generating and the effect of incorporating synthetic data is difficult to predict.
To understand these nuances further, language technology engineers often empirically investigate LLM-generated data with respect to improvements in downstream performance. 
In this light, we propose DKPS to provide the foundation for mathematical analysis yielding concrete statistical guarantees for the utility of 
the synthetic data outputs of transformer models and their utility in subsequent inference tasks.
Towards this goal, we herein present a framework for exploring the properties of synthetic data in the context of machine translation. 
Combining neural translation systems with word embeddings give tractable representations of the linguistic data that allow for fundamental properties like synthetic data bias and variance to be gainfully explored.
Beyond estimation, we also consider Data Kernel Perspective Space which allows us to visualize the different synthetic data models (batch versus sequential translations) under a variety of learning settings (here MLE and CPO).
DKPS provides a setting for exploring these black box models, and can be used as a preprocessing tool before statistical primitives are applied (e.g., testing, computing model distance, etc.).

Future directions for this work abound, as this is a first step in the direction of statistical synthetic data analysis.
Among the possible extensions are the following: with DKPS providing a framework for identifying systemic model bias, can we develop techniques which control or leverage that bias; 
considering (non)linear or local dimension reduction methods beyond PCA (see \cite{de2025low} for a bevy of alternate options);  
the effect of different language embedding systems \cite{enevoldsen2025mmteb}; 
exploring the interplay between higher dimensional representations and larger sample sizes in the analysis (which leads to the important question of optimal dimension selection); 
and possibly integrating DKPS into the CPO pipeline to learn the DKPS representations directly; among myriad other possibilities.
In future work, we will further investigate how DKPS performance guarantees can elucidate performance properties of subsequent inference tasks
for various applications, 
such as the development of machine translation models trained via back translation.

\bibliographystyle{IEEEtran}

\bibliography{IEEEabrv,ref}
\vspace{-5mm}

\begin{IEEEbiography}[{\includegraphics[width=1in,height=1.25in,clip,keepaspectratio]{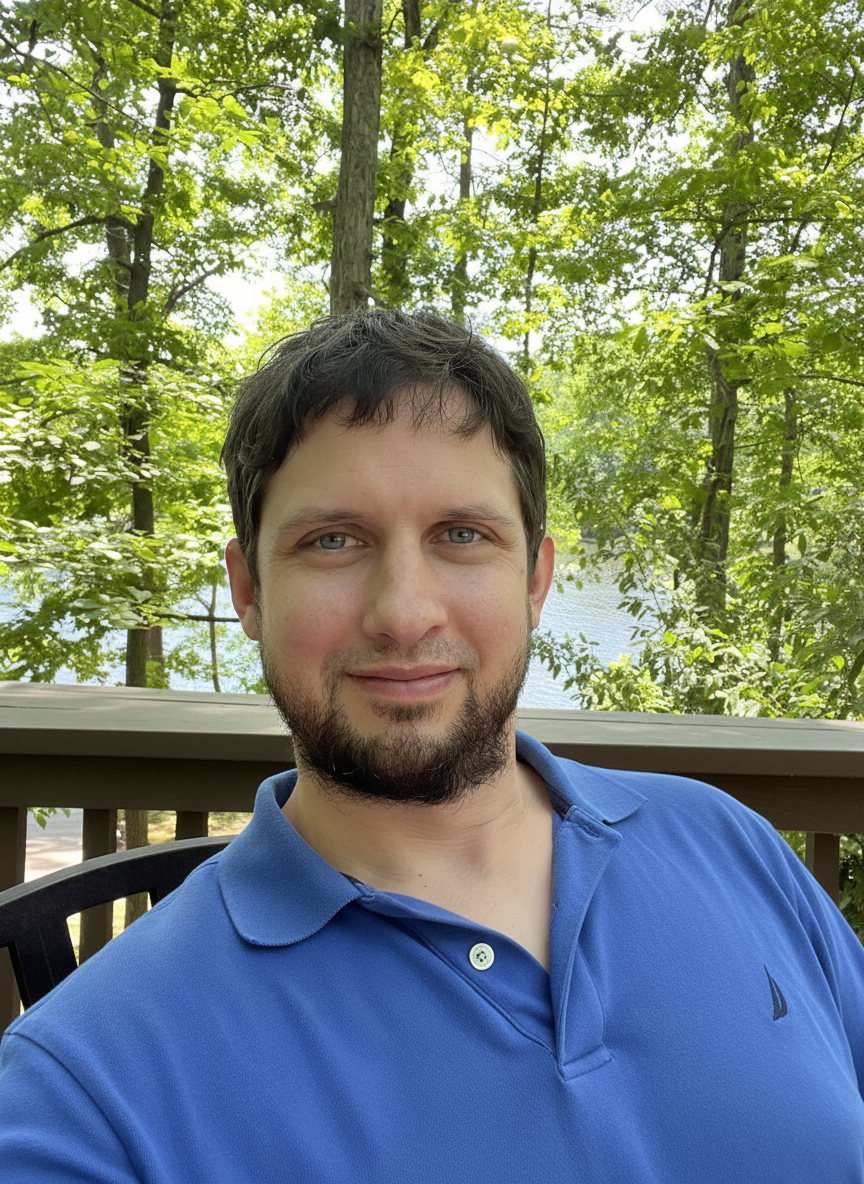}}]{Michael Browder} received his bachelor's degree from Duke University and is a graduate student in applied statistics at the University of Maryland, College Park.  He has been a supervisory mathematical statistician with the U.S. federal government since 2024.  His interests include statistics, optimization, and reinforcement learning, with applications to large language models and to the decision making of agents in interactive virtual environments.

\end{IEEEbiography}
\vspace{-5mm}

\begin{IEEEbiography}[{\includegraphics[width=1in,height=1.25in,clip,keepaspectratio]{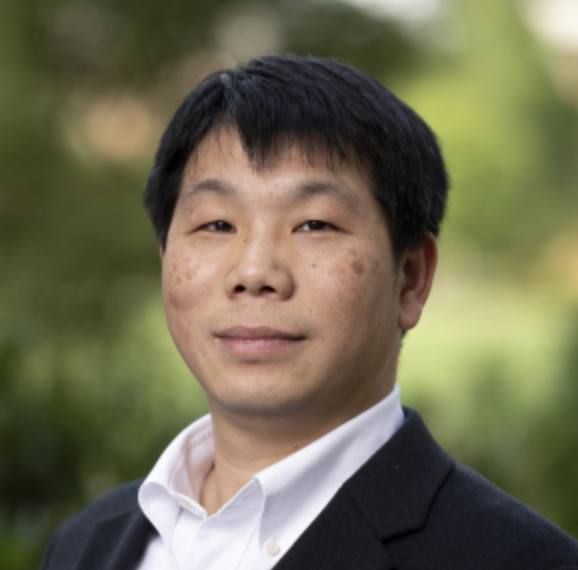}}]{Kevin Duh} received the B.S. from Rice University in 2003 and the PhD degree from the University of Washington in 2009, both in Electrical Engineering. From 2009 to 2012, he was a research associate at NTT; from 2012 to 2015, he was a assistant professor at the Nara Institute of Science and Technology (Japan). Since 2015, he is a senior research scientist at the Johns Hopkins University Human Language Technology Center of Excellence. His research focuses on natural language processing. 

\end{IEEEbiography}

\vspace{-5mm}
\begin{IEEEbiography}
[{\includegraphics[width=1in,height=1.25in,clip,keepaspectratio]{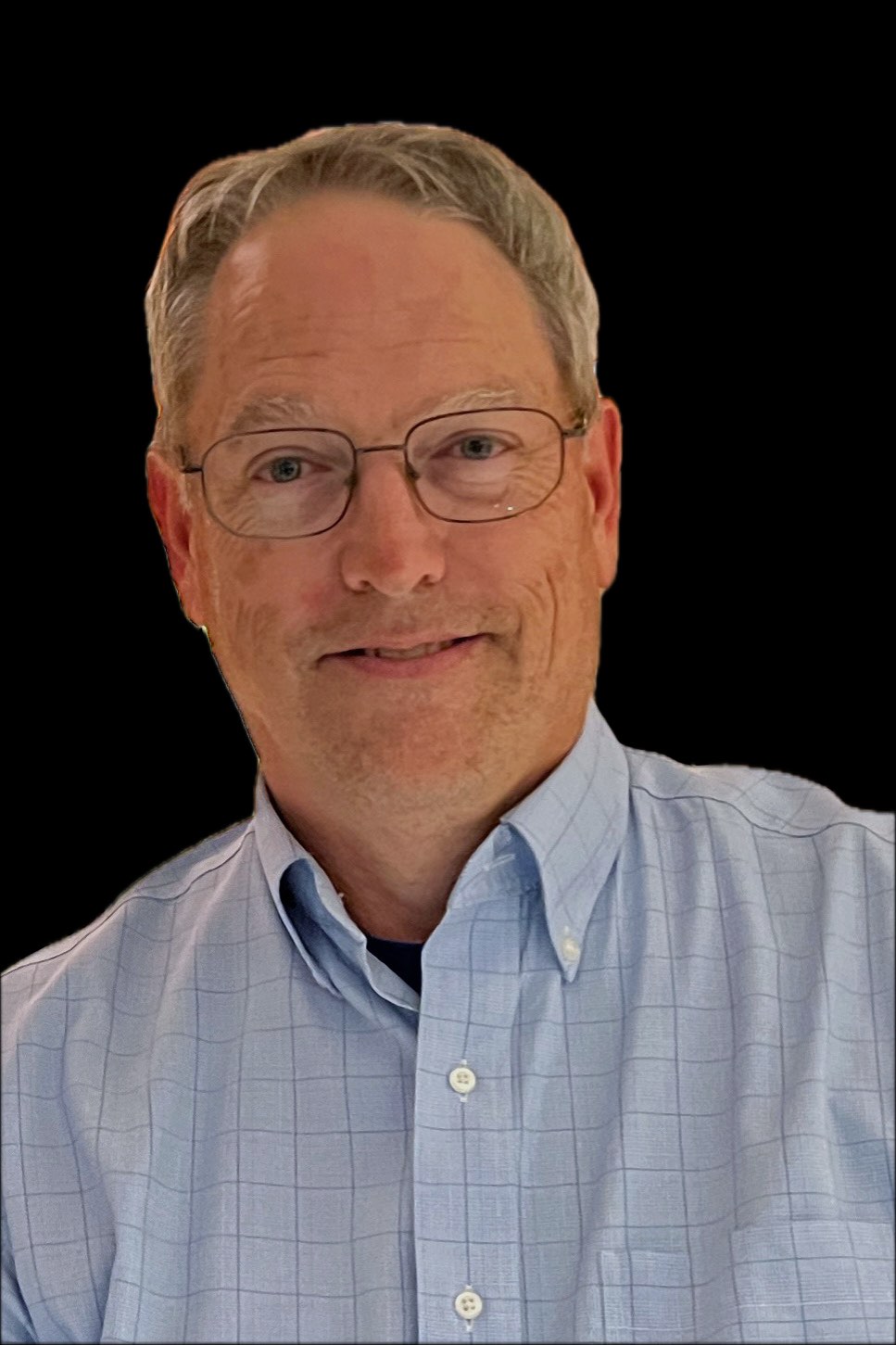}}]{J. David Harris} joined the Human Language Technology Center of Excellence (HLTCOE) at Johns Hopkins University after retiring from the U.S Government. In the early years of his public service, he was a trailblazer in applying data mining and machine learning techniques to speech and language processing. In 2004, he reached the Senior Executive Service level, as a Defense Intelligence Senior Leader, and from there, he oversaw various projects and organizations spanning research, technology, operations, and human capital management. Notably, he served as the first Director of the Laboratory for Analytic Sciences, where he led a collaborative effort involving government, academic, and industry partners to develop and implement technology and methods to meet intelligence analysis needs. Mr. Harris’s distinguished 32-year career earned him numerous awards and commendations, with a notable highlight being the Meritorious Presidential Rank Award. This prestigious award recognized his sustained professional, technical, and scientific accomplishments on behalf of the U.S. Government.    
\end{IEEEbiography}

\begin{IEEEbiography}[{\includegraphics[width=1in,height=1.25in,clip,keepaspectratio]{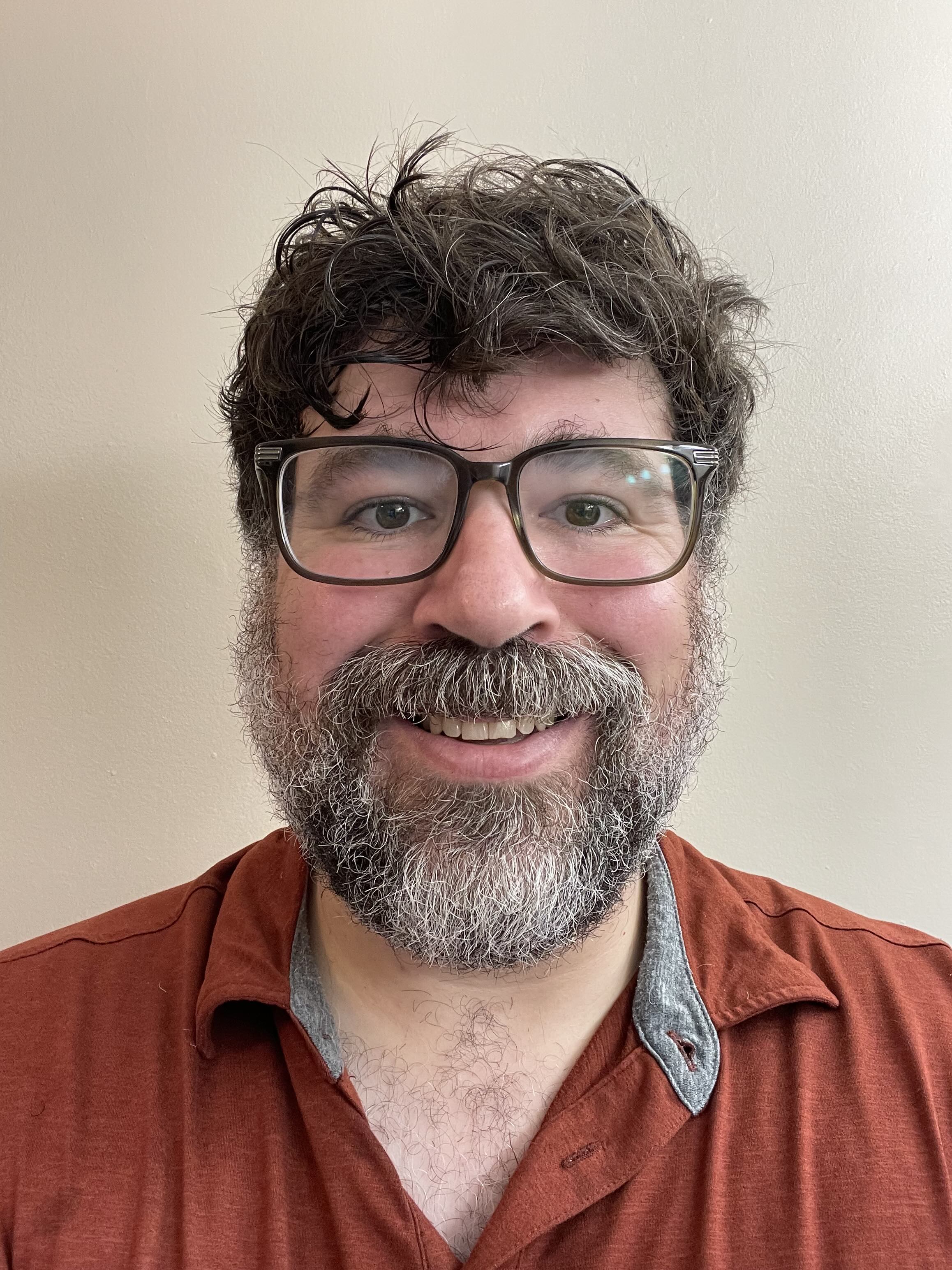}}]{Vince Lyzinski}
 received the BSc degree in mathematics from the University of Notre Dame, in
2006, the MA degree in mathematics from Johns
Hopkins University (JHU), in 2007, the MSE
degree in applied mathematics and statistics
from JHU, in 2010, and the PhD degree in applied
mathematics and statistics from JHU, in 2013.
From 2013-2014 he was a postdoctoral fellow
with the Applied Mathematics and Statistics
(AMS) Department, JHU. During 2014-2017, he
was a senior research scientist with the JHU
HLTCOE and an assistant research professor with the AMS Department, JHU.
From 2017-2019, he was on the Faculty in the Department of Mathematics
and Statistics at the University of Massachusetts Amherst. Since 2019 he has been on the Faculty in the Department of Mathematics at the University of Maryland, College Park, where he is currently a Professor. His research interests include graph matching, statistical
inference on random graphs, pattern recognition, dimensionality reduction, stochastic processes, and high-dimensional data analysis.
\end{IEEEbiography}

\begin{IEEEbiography}[{\includegraphics[width=1in,height=1.25in,clip,keepaspectratio]{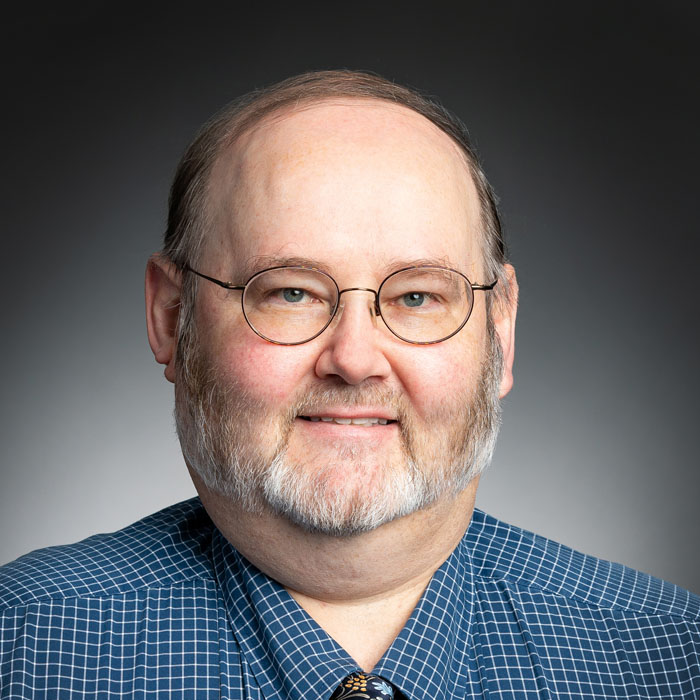}}]{Paul McNamee}
 received BS (electrical engineering) and MS (computer science) degrees from Johns Hopkins University, in 1990 and 1996, respectively. In 2008 he received the PhD degree in computer science from the University of Maryland Baltimore County.
 He holds longtime appointments with the Johns Hopkins University Applied Physics Laboratory and the Johns Hopkins Human Language Technology Center of Excellence.  His research interests have focused on multilingual text processing.
\end{IEEEbiography}

\begin{IEEEbiography}
[{\includegraphics[width=1in,height=1.25in,clip,keepaspectratio]{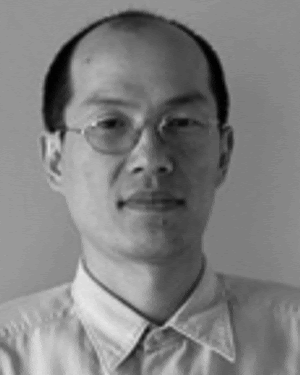}}]{Youngser Park} received the B.E. degree in electrical engineering from Inha University in Seoul, Korea in 1985, the M.S. and Ph.D. degrees in computer science from The George Washington University in 1991 and 2011 respectively. From 1998 to 2000 he worked at the Johns Hopkins Medical Institutes as a senior research engineer. From 2003 until 2011 he worked as a senior research analyst, and has been an associate research scientist since 2011 then research scientist since 2019 in the Center for Imaging Science at the Johns Hopkins University. At Johns Hopkins, he holds joint appointments in the Mathematical Institute for Data Science and the Human Language Technology Center of Excellence. His current research interests are clustering algorithms, pattern classification, and data mining for high-dimensional and graph data.
\end{IEEEbiography}

\begin{IEEEbiography}
[{\includegraphics[width=1in,height=1.25in,clip,keepaspectratio]{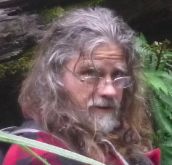}}]{Carey E. Priebe} received the BS degree in mathematics from Purdue University in 1984, the MS degree in computer science from San Diego State University in 1988, and the PhD degree in information technology (computational statistics) from George Mason University in 1993. From 1985 to 1994 he worked as a mathematician and scientist in the US Navy research and development laboratory system. Since 1994 he has been a professor in the Department of Applied Mathematics and Statistics at Johns Hopkins University. His research interests include computational statistics, kernel and mixture estimates, statistical pattern recognition, model selection, and statistical inference for high-dimensional and graph data. He is a Senior Member of the IEEE, an Elected Member of the International Statistical Institute, a Fellow of the Institute of Mathematical Statistics, and a Fellow of the American Statistical Association.
\end{IEEEbiography}

\begin{IEEEbiography}
[{\includegraphics[width=1in,height=1.25in,clip,keepaspectratio]{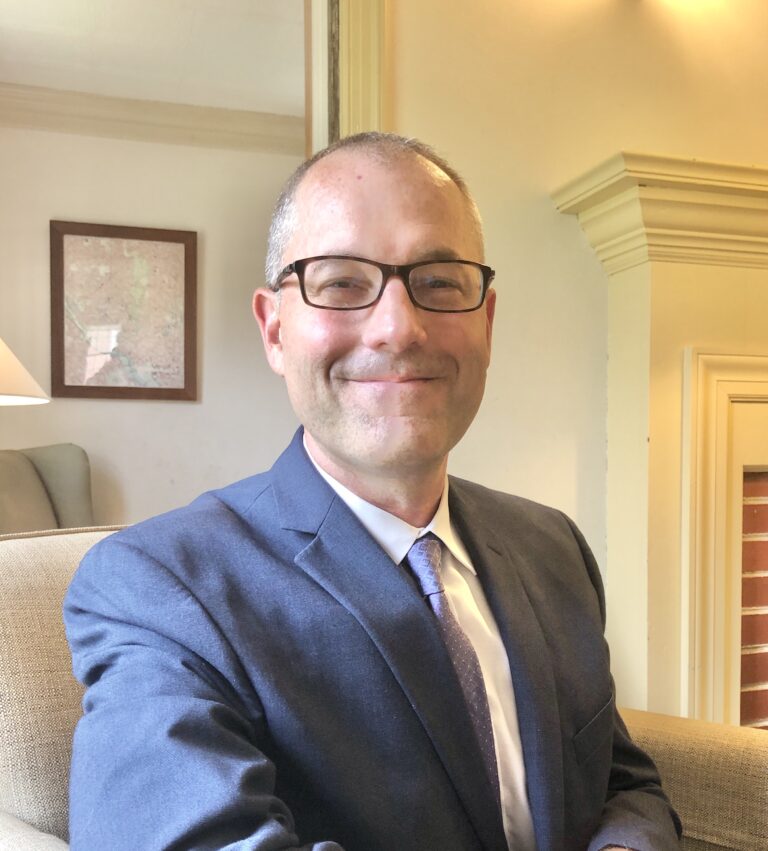}}]{Peter Viechnicki} received the BA degree in Classics and Linguistics from Princeton University in 1992, and the PhD in Linguistics from the University of Chicago in 2002.  After a career in government and the management consulting industry he was recruited to direct the Human Language Technology Center of Excellence at the Whiting School of Engineering at Johns Hopkins University from 2020 to 2025. His research interests include computational linguistics, phonetics, and cognitive science.    
\end{IEEEbiography}

\newpage
\onecolumn
\appendix

\subsection{Mahalanobis Distance computations}
\label{sec:dkps}
In the DKPS CPO setting, the Mahalanobis distances are computed as follows:  consider a single sentence with $x$ denoting the true (human) translation, and $\{y^{(x)}_{w,i}\}_{i=1}^{10}$ the sequential translations and $\{y^{(x)}_{l,i}\}_{i=1}^{10}$ the batch translations.  Let $\Sigma^{(x)}_w$ and $\Sigma^{(x)}_l$ denoted the respective fitted covariances.
The distances for the DKPS of the sequential embeddings are computed via:
$$
D_x(y^{(x)}_{w,i},y^{(x)}_{w,j})=\sqrt{ (y^{(x)}_{w,i}-y^{(x)}_{w,j})^T(\Sigma^{(x)}_w)^{-1}(y^{(x)}_{w,i}-y^{(x)}_{w,j})};\quad
D_x(x,y^{(x)}_{w,j})=\sqrt{ (x-y^{(x)}_{w,j})^T(\Sigma^{(x)}_w)^{-1}(x-y^{(x)}_{w,j})};
$$
for the DKPS of the batch embeddings are computed via:
$$
D_x(y^{(x)}_{l,i},y^{(x)}_{l,j})=\sqrt{ (y^{(x)}_{l,i}-y^{(x)}_{l,j})^T(\Sigma^{(x)}_l)^{-1}(y^{(x)}_{l,i}-y^{(x)}_{l,j})};\quad
D_x(x,y^{(x)}_{l,j})=\sqrt{ (x-y^{(x)}_{l,j})^T(\Sigma^{(x)}_l)^{-1}(x-y^{(x)}_{l,j})};
$$
and for the joint DKPS embeddings
\begin{align*}
   D_x(y^{(x)}_{l,i},y^{(x)}_{l,j})&=\sqrt{ (y^{(x)}_{l,i}-y^{(x)}_{l,j})^T(\Sigma^{(x)}_l)^{-1}(y^{(x)}_{l,i}-y^{(x)}_{l,j})};\quad
D_x(x,y^{(x)}_{l,j})=\sqrt{ (x-y^{(x)}_{l,j})^T(\Sigma^{(x)}_l)^{-1}(x-y^{(x)}_{l,j})};\\
D_x(y^{(x)}_{w,i},y^{(x)}_{w,j})&=\sqrt{ (y^{(x)}_{w,i}-y^{(x)}_{w,j})^T(\Sigma^{(x)}_w)^{-1}(y^{(x)}_{w,i}-y^{(x)}_{w,j})};\quad
D_x(x,y^{(x)}_{w,j})=\sqrt{ (x-y^{(x)}_{w,j})^T(\Sigma^{(x)}_w)^{-1}(x-y^{(x)}_{w,j})};\\
D_x(y^{(x)}_{w,i},y^{(x)}_{l,j})&=\max\left(\sqrt{ (y^{(x)}_{w,i}-y^{(x)}_{l,j})^T(\Sigma^{(x)}_w)^{-1}(y^{(x)}_{w,i}-y^{(x)}_{l,j})},\sqrt{ (y^{(x)}_{w,i}-y^{(x)}_{l,j})^T(\Sigma^{(x)}_l)^{-1}(y^{(x)}_{w,i}-y^{(x)}_{l,j})}\right)
\end{align*}
Adapting the notation of Section \ref{sec:bVs} and trying to mimic the computation $D$ in Section \ref{sec:DKPS} in this new setting, we compute 
$D$ between any pair of translation models $f^{(i)},f^{(j)}$ as follows:
In the sequential setting (the batch setting being analogous)
$$D(f^{(i)},f^{(j)})=\begin{cases}
       \frac{1}{m}\sqrt{\sum_x D_x(y^{(x)}_{l,i},y^{(x)}_{l,j})^2}&\text{ if }i,j>1;\\
        \frac{1}{m}\sqrt{\sum_x D_x(y^{(x)}_{l,i},x)^2}&\text{ if }i>1,j=1;
\end{cases}
$$
In the joint setting, we have 
$$
D(f^{(i)},f^{(j)})=\begin{cases}
\frac{1}{m}\sqrt{\sum_x D_x(y^{(x)}_{w,i-1},y^{(x)}_{l,j-10})^2}&\text{ if }i\in[2,11], j\in[12,21]\\
\frac{1}{m}\sqrt{\sum_x D_x(y^{(x)}_{w,i-1},y^{(x)}_{w,j-1})^2}&\text{ if }i\in[2,11], j\in[2,11]\\
\frac{1}{m}\sqrt{\sum_x D_x(y^{(x)}_{l,i-10},y^{(x)}_{l,j-10})^2}&\text{ if }i\in[12,21], j\in[12,21]\\
\frac{1}{m}\sqrt{\sum_x D_x(x,y^{(x)}_{l,j-10})^2}&\text{ if }i=1, j\in[12,21]\\
\frac{1}{m}\sqrt{\sum_x D_x(y^{(x)}_{w,i-1},x)^2}&\text{ if }i\in[2,11], j=1.\\
\end{cases}
$$
\newpage
\subsection{Additional Figures}
\label{sec:figs}

\begin{figure}[h!]
  \centering
\includegraphics[width=0.4\textwidth]{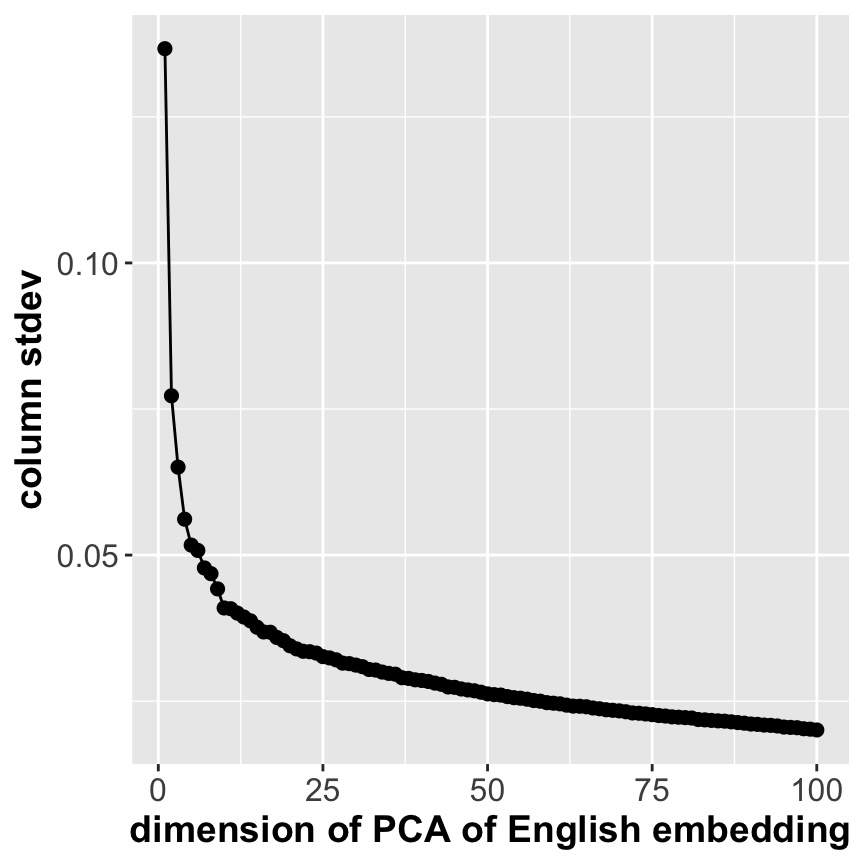}
\includegraphics[width=0.4\textwidth]{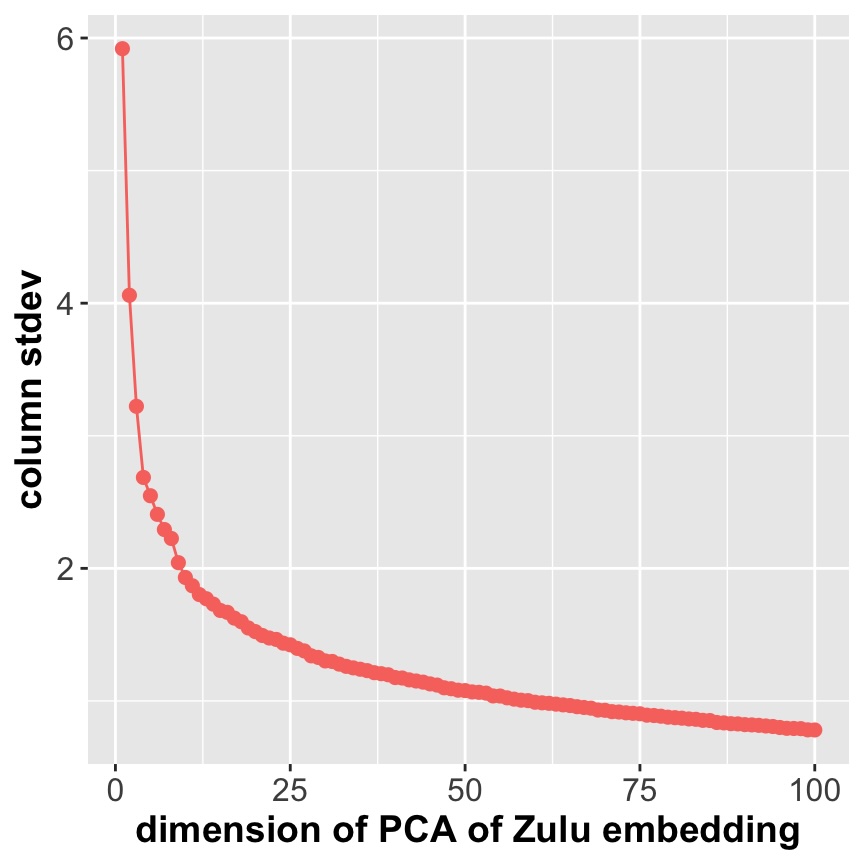}\\
\includegraphics[width=0.4\textwidth]{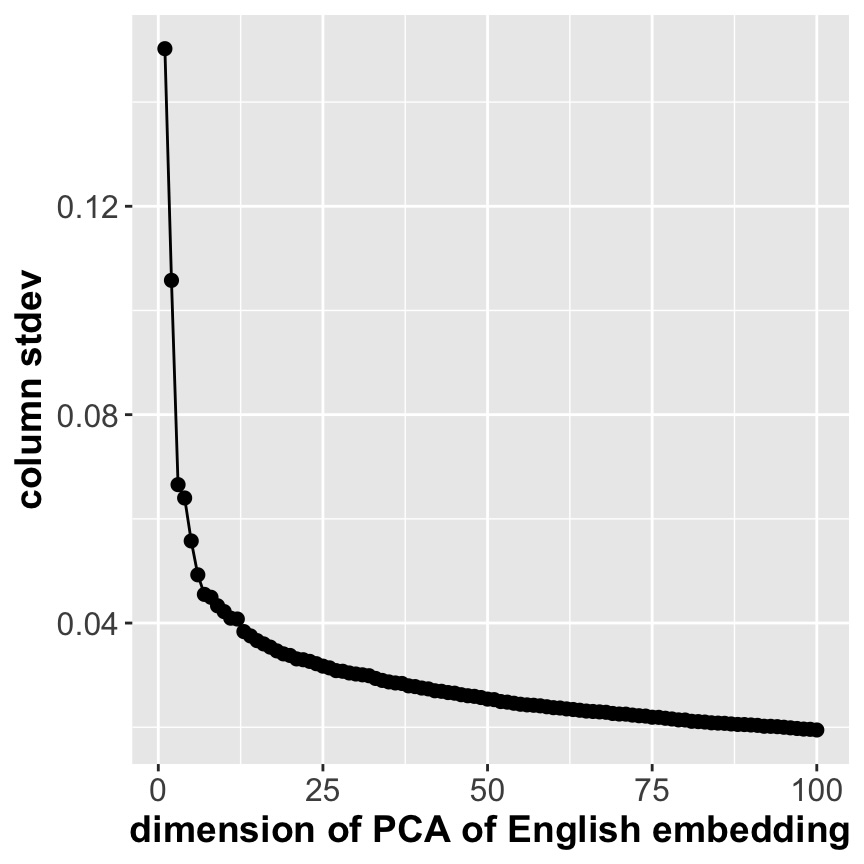}
\includegraphics[width=0.4\textwidth]{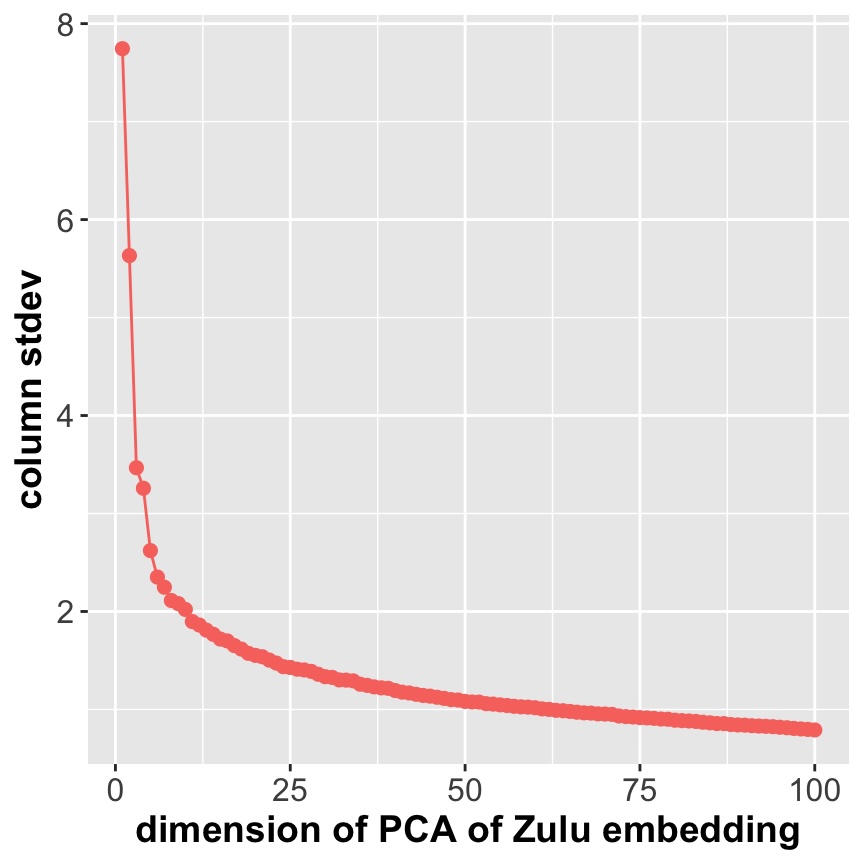}
  \caption{
Principal component scree plots for English embeddings (L) and human-produced Zulu embeddings (R).  In sample is top row, out-of-sample bottom.
}
\label{fig:scree}
\end{figure}

\begin{figure}[t!]
  \centering
\includegraphics[width=0.8\textwidth]{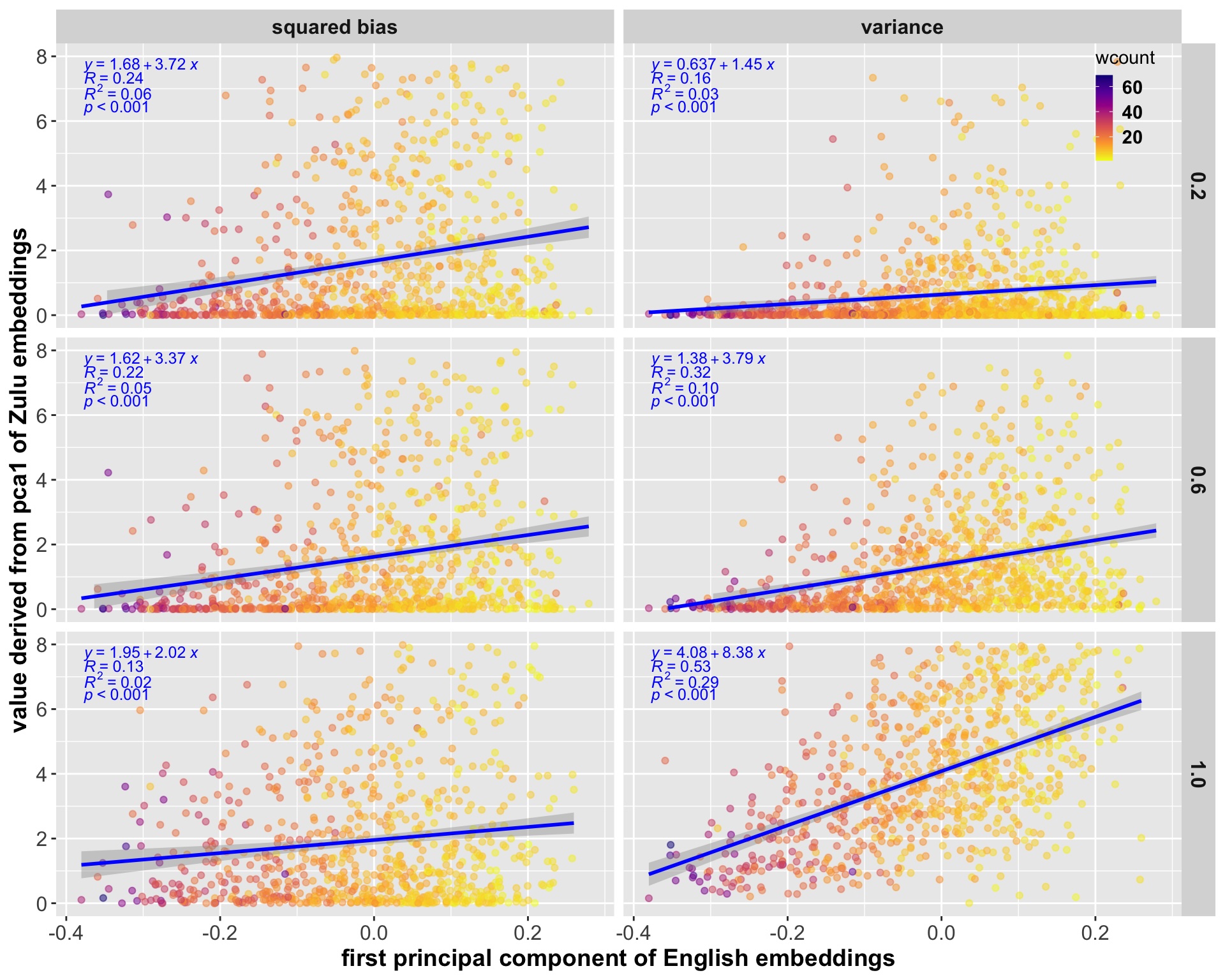}
  \caption{
For temperate $t=0.2,$ (top), 0.6 (middle), and 1 (bottom), we plot: (Left) Squared bias for the 10 batched synthetic translations with respect to the human translation; (Right) Estimated variance for the 10 synthetic translations. 
Points are colored by the number of words of English sentences.
}
\label{fig:bias_variance}
\end{figure}

\begin{figure}[t!]
  \centering
\includegraphics[width=.8\textwidth]{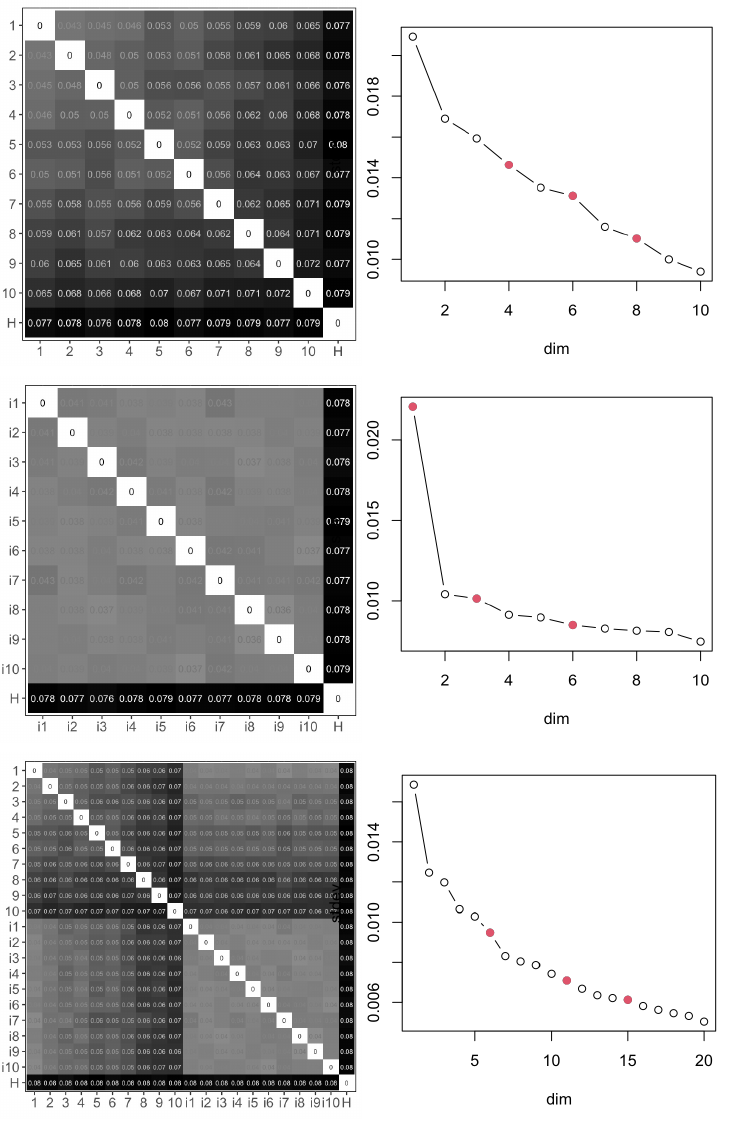}
  \caption{
In the in-sample setting of Section \ref{sec:bVs}, we plot the distance matrices in 1-dimensional PCA space and MDS SCREE plots (detected elbows colored in red) between the human translation and the 
batched translations (T), the
sequential translations (M), and the combined batched and sequential translations (B).
}
\label{fig:Din}
\end{figure}

\begin{figure}[t!]
  \centering
\includegraphics[width=.8\textwidth]{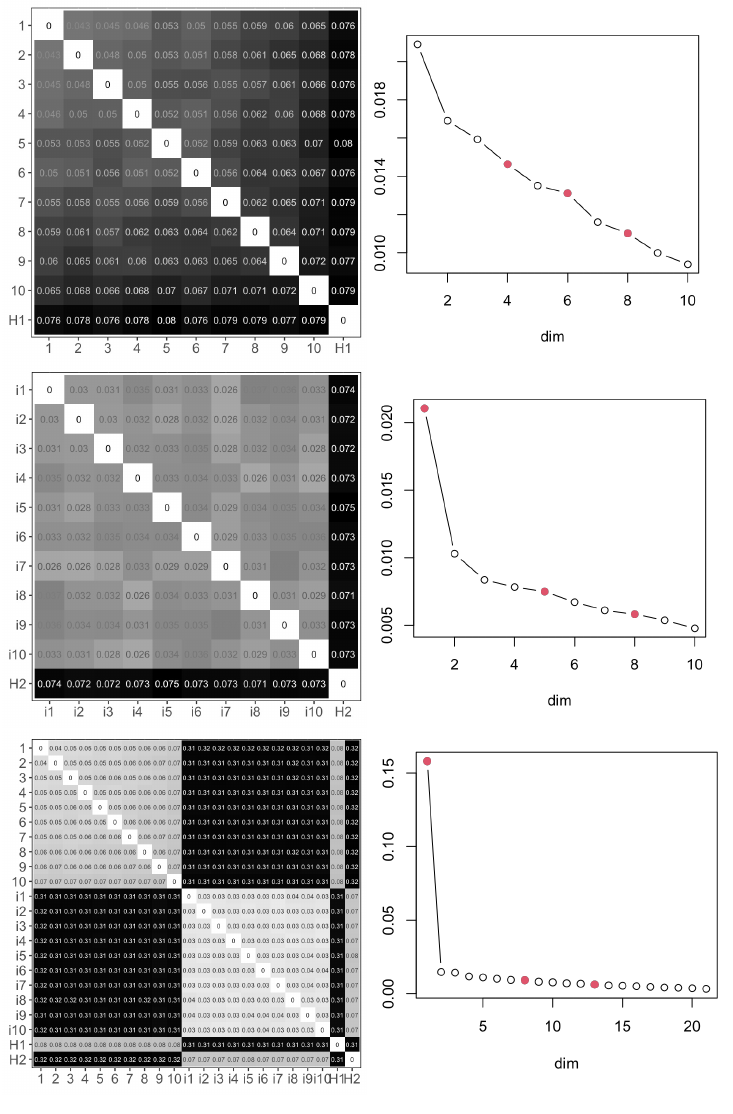}
  \caption{
In the in-sample and OOS setting of Section \ref{sec:bVs}, we plot the distance matrices in 1-dimensional PCA space and MDS SCREE plots (detected elbows colored in red) between the human translation and the 
batched translations (T), the
sequential OOS translations (M), and the combined batched and sequential translations (B).
The bottom row shows SCREE plots for each distance matrix along with automated detected elbows in the SCREE (colored in red).
}
\label{fig:Dout}
\end{figure}

\end{document}